\documentclass{article} % For LaTeX2e
\usepackage{iclr2026_conference, times}

\usepackage{amsmath,amsfonts,bm}

\def\eqref#1{equation~\ref{#1}}
\def\1{\bm{1}}

\DeclareMathAlphabet{\mathsfit}{\encodingdefault}{\sfdefault}{m}{sl}
\SetMathAlphabet{\mathsfit}{bold}{\encodingdefault}{\sfdefault}{bx}{n}

\usepackage{hyperref}
\usepackage{url}
\usepackage{graphicx}
\usepackage{enumitem}
\usepackage{tabularx}
\usepackage{float}
\usepackage{ulem}
\usepackage{pifont}
\usepackage{adjustbox}
\usepackage{amsfonts}
\usepackage{booktabs}
\usepackage{multirow}
\usepackage[T1]{fontenc}
\usepackage{wrapfig}
\usepackage{algorithm}
\usepackage{algorithmic}

\title{ProtoTS: Learning Hierarchical Prototypes \\ for Explainable Time Series Forecasting}

\author{Ziheng~Peng$^{1,2,3}$\thanks{Equal contribution. Work done during internship at DAMO Academy.} \quad
Shijie~Ren$^{1,2,3}$\footnotemark[1] \quad 
Xinyue~Gu$^{2,3}$\\
\textbf{Linxiao~Yang}$^{2,3}$\quad
\textbf{Xiting~Wang}$^{1}$\thanks{Corresponding author.}\,
\thanks{Work partly done at Beijing Key Laboratory of Research on Large Models and Intelligent Governance and Engineering
Research Center of Next-Generation Intelligent Search and Recommendation, MOE.} \quad
\textbf{Liang~Sun}$^{2,3}$\footnotemark[2] \\
$^{1}$Gaoling School of Artificial Intelligence, Renmin University of China Beijing \quad \\
$^{2}$DAMO Academy, Alibaba Group \quad
$^{3}$Hupan Laboratory \\
\texttt{\{ziheng.peng, shj\_ren, xitingwang\}@ruc.edu.cn} \\
\texttt{\{guxinyue.gxy, linxiao.ylx, liang.sun\}@alibaba-inc.com}
}

\iclrfinalcopy % Uncomment for camera-ready version, but NOT for submission.
\begin{document}

\maketitle

\begin{abstract}
While deep learning has achieved impressive performance in time series forecasting, it becomes increasingly crucial to understand its decision-making process for building trust in high-stakes scenarios. Existing interpretable models often provide only local and partial explanations, lacking the capability to reveal how heterogeneous and interacting input variables jointly shape the overall temporal patterns in the forecast curve. We propose ProtoTS, a novel interpretable forecasting framework that achieves both high accuracy and transparent decision-making through modeling prototypical temporal patterns. ProtoTS computes instance-prototype similarity based on a denoised representation that preserves abundant heterogeneous information. The prototypes are organized hierarchically to capture global temporal patterns with coarse prototypes while capturing finer-grained local variations with detailed prototypes, enabling expert steering and multi-level interpretability. Experiments on multiple realistic benchmarks, including a newly released LOF dataset, show that ProtoTS not only exceeds existing methods in forecast accuracy but also delivers expert-steerable interpretations for better model understanding and decision support. The source code is available at~\url{https://github.com/SKURA502/ProtoTS}.
\end{abstract}

\section{Introduction}

Time series forecasting has been widely applied in high-stakes scenarios such as load forecasting~\citep{jiang2024interpretable, yang2023interactive}, energy management~\citep{deb2017review, weron2014electricity}, weather prediction~\citep{angryk2020multivariate, karevan2020transductive}, all of which involve considerable financial impacts. In these applications, while achieving high forecast accuracy is crucial, understanding why and how the model makes specific predictions is equally important. It aids in preventing substantial financial losses and building the trust necessary~\citep{rojat2021explainable}.

\begin{figure}
    \centering
    \includegraphics[width=1\linewidth]{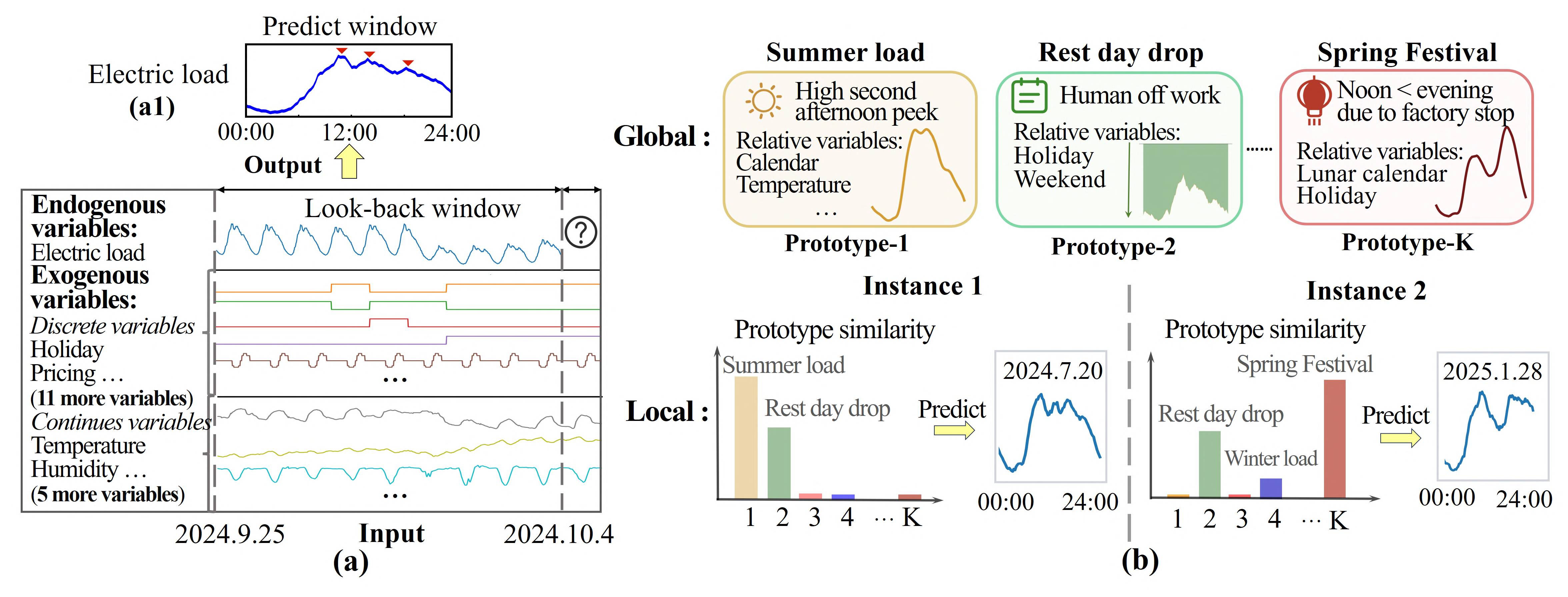}
    \vspace{-2em}
    \caption{(a) An example of time series forecasting with numerous heterogeneous variables, where exogenous variables (e.g., temperature, holiday) influence the evolution of endogenous variables (e.g., electric load). (b) Illustration of prototypical explanation method: a set of learned prototypes provides a user-friendly global overview of typical temporal patterns. For each instance, model computes its similarity to all prototypes to form a prediction, enabling detailed local interpretation.}
    \label{figure1}
    \vspace{-0.5em}
\end{figure}

A range of explainable time series forecasting methods have been developed to simultaneously ensure interpretability and good predictive performance~\citep{oreshkin2019n, lim2021temporal, zhao2024disentangled, lin2024cyclenet}. However, their overall interpretability and potential for further performance improvement are limited, since they mainly provide local, partial explanations for both the output and input sides:
\begin{itemize}[nosep,leftmargin=1em,labelwidth=*,align=left]
\item \textbf{C1}: 
For the output side, existing methods~\citep{lim2021temporal, zhao2024disentangled} mainly explain the prediction at individual time steps, \textbf{lacking the ability to help users quickly interpret the reasons behind the overall trend in the forecast curve}. For example, they fail to explain why three peaks occur in the predicted electricity load curve  at the noon, afternoon, and night, respectively, and why the three peaks gradually descend (Figure~\ref{figure1}(a1)). 
Understanding such overall temporal patterns is of great importance in real-world scenarios. For example, power system experts need to ensure demand peaks are identified correctly, so that they can make dispatch decisions~\citep{6721569}, such as whether additional electricity should be purchased from external sources. Failing to provide such analyses limits interpretability and prevents experts from steering the model to improve its accuracy.% \looseness=-1
\item \textbf{C2}: For the input side, existing explanations are often limited to certain types of variables, such as focusing solely on endogenous variables~\citep{lin2024cyclenet}. However, as shown in Figure~\ref{figure1}(a), \textbf{there are many different types of input variables, and understanding how their interactions impact the temporal patterns is critical}. For example, the typical temporal pattern of extreme summer conditions can only be observed when both temperature and seasonal variables are considered together.
\end{itemize}

To address these challenges, we propose a \textbf{Proto}typical \textbf{T}ime \textbf{S}eries Forecasting framework (\textbf{ProtoTS}) that effectively models the overall temporal patterns (C1) and enables prototypes to capture the interactions among inputs (C2).

As shown in Figure~\ref{figure1}(b), each of our prototypes corresponds to a typical temporal pattern. For example, the curve of the Spring Festival prototype\footnote{The name ``Spring Festival'' is now manually summarized based on relevant covariates and temporal patterns, though it can also be generated using large language models.} indicates that there are peaks in the morning and evening during the Spring Festival period. This pattern occurs at specific combinations of covariates (i.e., when the inputs indicate both a holiday and the first month of the lunar calendar), enabling it to explain both the formation of the overall temporal pattern and how it is influenced by the combination of multiple input features (C2). 
These few prototypes provide a global explanation by summarizing typical patterns across the entire dataset, thereby supporting expert intervention (e.g., refining the temporal pattern of the prototype, Sec.~\ref{sec-casestudy}). At the same time, the model improves accuracy by effectively modeling the local interactions of covariates.

To construct an effective end-to-end framework, we introduce the following technical innovations:

\begin{itemize}[nosep,leftmargin=1em,labelwidth=*,align=left]
    \item \textbf{Hierarchical prototype learning strategy (C1)}. Traditional prototypes are typically limited to explaining individual classification results, while we extend the prototypes so that they model a temporal pattern, which consists of a sequence of regression results. Learning such a sequential temporal pattern is hard to balance predictive performance with interpretability. Using too few prototypes leads to poor forecasting accuracy, while employing too many can compromise interpretability by producing overly similar patterns. To address this, we design a hierarchical prototype learning strategy, where a small set of coarse prototypes ensures global interpretability, and prototypes are progressively refined into lower levels to capture local details and improve accuracy. This design allows experts to intervene by specifying which prototypes should be further split, enabling efficient and user-friendly model adjustments.
    \vspace{0.6em}
    \item \textbf{Multi-channel prototype similarity computing (C2)}. To enable effective modeling of the interactions among input variables of different types and their impact on the output curve patterns, we compute the prototype-instance similarity by incorporating a multi-channel embedding and bottleneck fusion mechanism, 
    which improves both the interpretability and the accuracy of the model.\looseness=-1
\end{itemize}

Our framework achieves state-of-the-art performance with good interpretability and steerability. In particular, ProtoTS reduces MSE by 48.3\% and MAE by 20.9\% compared to the state-of-the-art model on the LOF dataset. Moreover, its accuracy decreases much more slowly when the amount of training data is reduced, compared to the baselines. 
A case study demonstrates that our explanations enable an easy-to-understand overall and detailed understanding. Expert edits based on the explanations reduced MSE by 0.009 and further improved explainability.

\section{Related Work}

\textbf{Time Series Forecasting with Exogenous Variables}\quad Time series forecasting with exogenous variables has been extensively explored in both classical and modern approaches. Traditional statistical models such as ARIMAX~\citep{williams2001multivariate} and SARIMAX~\citep{vagropoulos2016comparison} extend the ARIMA framework to incorporate correlations between exogenous and endogenous variables. Seeking richer dynamics, Transformer families like TFT~\citep{lim2021temporal} and TimeXer~\citep{wang2024timexer}, introduce attention mechanisms to tackle this task. Parallel work in MLPs pursues efficiency and long-horizon stability: NBEATSx~\citep{oreshkin2019n} appends a dedicated stack for auxiliary factors on top of a residual basis expansion, TiDE~\citep{das2023long} factorizes temporal and feature projections with two dense encoders to harness known‑future variables, and TSMixer~\citep{ekambaram2023tsmixer, chen2023tsmixer} alternates mixing layers to effectively model their interactions. Outside deep learning, generic tabular learners such as XGBoost~\citep{chen2016xgboost}, LightGBM~\citep{ke2017lightgbm}, and GAM~\citep{yang2023interactive}, perform well when exogenous variables are informative and strongly correlated with the target. Our method, ProtoTS, achieves strong performance on this task while offering interpretability and steerability for domain experts, effectively handling large amounts of heterogeneous variables while filtering out irrelevant information.

% These models often perform well when informative exogenous variables are present, particularly in settings with strong feature-target dependencies.
%, and CITRAS~\citep{yamaguchi2025citras} refines cross-variate attention with KV Shift and score smoothing

\textbf{Time Series Interpretability}\quad Time series interpretability can be divided into post-hoc and ante-hoc approaches. Post-hoc interpretability focuses on unveiling the reasoning of pre-existing black-box models~\citep{gu2025explainable}. Generic explanation techniques such as SHAP~\citep{lundberg2017unified}, LIME~\citep{ribeiro2016should}, IG~\citep{sundararajan2017axiomatic} can be directly applied to time-series models. \citet{crabbe2021explaining} learns sparse perturbation masks over the time-feature dimensions, struggling to quantify each input’s effect on a regression forecast. Consequently, regression tasks tend to favour ante-hoc models whose interpretability is built in by design. Following the attribution-centric view of post‑hoc explanations, TFT~\citep{lim2021temporal} provides explorable attention over input time steps, and DiPE-Linear~\citep{zhao2024disentangled} parameterizes filters in both temporal and spectral domains to visualise the regions of interest. N-BEATS~\citep{oreshkin2019n} uses residual blocks for generic, trend, and seasonal components, and \citet{olivares2023neural} extends the scheme to exogenous variables. These works are limited to local explanations and fail to provide a global view that experts need for a shared understanding. CycleNet~\citep{lin2024cyclenet} identifies common periodic patterns in time-series data but only focuses on endogenous variables. In contrast, ProtoTS holds a global set of prototypical patterns and explains each forecast by matching it to them, offering both global overview and local interpretation. Its hierarchical structure supports expert steering at multiple levels, ensuring faithful overall understanding.

\textbf{Prototypical Time Series Model}\quad Previous works have already shown the value of prototypes in time series models. These methods typically fall into two categories. 1) prototypes decoded as intermediate representations~\citep{shen2025learn, li2023prototype, jin2023time} or model parameters~\citep{chen2024similarity}. For example, \citet{jin2023time} uses prototypes to map time-series embeddings into textual embeddings for large language models. Such designs enrich the model’s architecture but leave the prototypes detached from the output. 2) prototypes directly mapped to output variables~\citep{queen2023encoding, liu2024timex++, ming2019interpretable, obermair2023example}, serving as typical examples of the outputs. While the latter offers stronger interpretability, existing approaches decode a prototype into only a single output variable (e.g., a class prediction). In contrast, our method is, to the best of our knowledge, the first to decode a prototype into a sequence of output variables (e.g., a 96-step forecasting curve), enabling holistic interpretation of temporal dynamics. 

\section{Method}

\label{method}
\begin{figure}[t]
    \centering
    \includegraphics[width=1\linewidth]{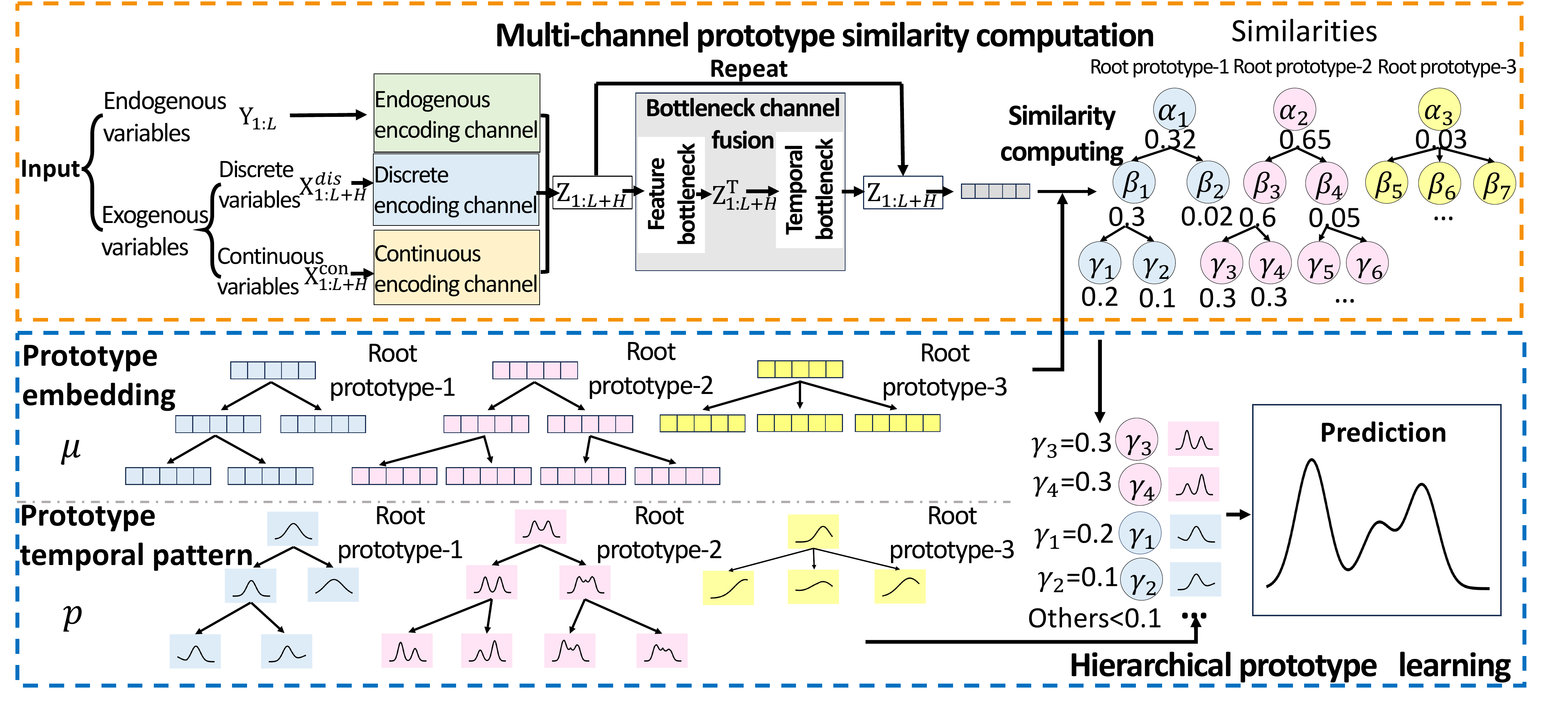}
    \vspace{-1.5em}
    \caption{The overall framework of ProtoTS, which comprises two main modules: the multi-channel prototype similarity computation module and the hierarchical prototype learning module.}
    \vspace{-1em}
    \label{figure2}
\end{figure}

%\subsection{Problem Formulation and Framework Overview} %\quad 
\subsection{Formulation and Framework}

\textbf{Problem Formulation}\quad 
We consider the task of time series forecasting with exogenous variables, as shown in Figure~\ref{figure1}(a), since abundant heterogeneous exogenous variables provide rich information to boost endogenous predictability. Formally, let the historical endogenous variables (e.g., electric load) as $\mathbf{Y}_{1:L} = \{\mathbf{y}_1, \mathbf{y}_2, \dots, \mathbf{y}_L\} \in \mathbb{R}^{L \times 1}$ and the associated exogenous variables (e.g., temperature and holiday) as $\mathbf{X}_{1:L} = \{\mathbf{x}_1, \mathbf{x}_2, \dots, \mathbf{x}_L\} \in \mathbb{R}^{L \times C}$, where $L$ is the look-back window size and $C$ is the number of exogenous variables. In many practical scenarios~\citep{wessel2020using, lin2023prediction}, future exogenous variables $\mathbf{X}_{L+1:L+H} = \{\mathbf{x}_{L+1}, \dots, \mathbf{x}_{L+H}\} \in \mathbb{R}^{H \times C}$ are also available. For example, in the LOF dataset, we obtain forecasted weather variables (e.g., temperature) from commercial weather services, and calendar variables (e.g., holiday flags) are accessible in advance. In our setting, we explicitly incorporate future exogenous variables, as these substantially improve forecasting performance. The goal is to predict the future curve of the endogenous variable for the next $H$ time steps, denoted as $\mathbf{Y}_{L+1:L+H} = \{\mathbf{y}_{L+1}, \dots, \mathbf{y}_{L+H}\} \in \mathbb{R}^{H \times 1}$. The forecasting model $\mathcal{F}$ takes the input variables to predict the future endogenous variables $\widehat{\mathbf{Y}}$:
\begin{equation}
\widehat{\mathbf{Y}}_{L+1:L+H} = \mathcal{F} \Big(\mathbf{Y}_{1:L}, \mathbf{X}_{1:L}, \mathbf{X}_{L+1:L+H}\Big). 
\end{equation}
 
\textbf{Framework Overview}\quad 
The \textbf{ProtoTS} framework is illustrated in Figure~\ref{figure2}. It contains two modules. The \textbf{multi-channel prototype similarity computation} module
effectively models the interactions between input variables and their impact on prototypes.  The \textbf{hierarchical prototype learning} module 
structures prototypes hierarchically, with coarse-grained root prototypes capturing global patterns and fine-grained child prototypes modeling local variations, ensuring good explainability, steerability, and accuracy. The two modules will be explained in detail in Sections~\ref{sec:multi-channel} and~\ref{sec:hierarchical}.

\subsection{Multi-Channel Prototype Similarity Computation}
\label{sec:multi-channel}

This module ensures effective modeling of interactions between input variables and their relations with prototypes by using three steps: the \textbf{multi-channel embedding} step  effectively embeds heterogeneous variables in separate channels, the \textbf{bottleneck channel fusion} step then models complex interactions between variables without causing overfitting and outputs the final input embedding, and the \textbf{prototype similarity computing} step finally estimates the relations between each input and the prototypes by computing the similarities between their embeddings.  
%The resulting input embedding is then used to compute the similarity between this input and the prototype,  
%(\textbf{multi-channel embedding}) 
%filters noise by using bottleneck fusion. This 
%It then computes similarities to a set of learned prototypes, forming the prediction as a linear combination of prototypical temporal patterns. These prototypes are structured hierarchically, with coarse-grained root prototypes capturing global patterns and fine-grained child prototypes modeling local variations.

\textbf{Multi-Channel Embedding}\quad 
In ProtoTS, we adopt a multi-channel embedding where endogenous and exogenous variables are processed separately to achieve optimal encoding for each type. Specifically, endogenous variables are directly encoded through an \textit{endogenous encoding channel}, implemented as a non-linear projection $\gamma$ using a multi-layer perceptron with activation functions, a widely used design~\citep{das2023long}. For exogenous variables, we follow standard practice in tabular learning~\citep{wang2022transtab}. We categorize them into discrete variables (e.g., is holiday, day of week) and continuous variables (e.g., temperature, humidity), and apply tailored embedding methods to best capture the properties of each variable. Concretely, exogenous variables $\mathbf{x}_t$ at time $t$ are decomposed into $\mathbf{x}_t^{\text{dis}} \in \mathbb{N}^{C_{\text{dis}}}$ and $\mathbf{x}^{\text{con}}_t \in \mathbb{R}^{C_{\text{con}}}$, $C_{\text{dis}}$ and $C_{\text{con}}$ refer to the number of discrete and continuous variables. Discrete variable $\mathbf{x}_{t,j}^{\text{dis}}$ are processed through a \textit{discrete encoding channel} using dedicated embedding tables $\mathbf{E}_j \in \mathbb{R}^{|\Omega_j| \times d}$, where $\Omega_j$ is the vocabulary and $d$ is the embedding dimension. Continuous variables $\mathbf{x}^{\text{con}}_{t,j}$ are projected into the embedding space through a \textit{continuous encoding channel}, applying variable-specific non-linear projections $\psi_j$. This multi-channel embedding allows the model to better preserve the information contained in heterogeneous variables~\cite{munz2024exploring}.

\textbf{Bottleneck Channel Fusion}\quad To effectively fuse all variables information at time step $t$ and obtain a full representation, we aggregate the embeddings of all variables through addition, which integrates information across heterogeneous variables without additional parameters~\citep{arora2017simple}:
\begin{equation}
    \mathbf{Z}_t = 
    \begin{cases}
        \gamma(\mathbf{y}_t) + \sum\limits_{j = 1}^{C_{\text{dis}}} \mathbf{E}_j(\mathbf{x}^{\text{dis}}_{t,j}) + \sum\limits_{j = 1}^{C_{\text{con}}} \psi_j(\mathbf{x}^{\text{con}}_{t,j}), & t \in [1 : L] \quad \text{ (Look-back Window)} \\
        \sum\limits_{j = 1}^{C_{\text{dis}}} \mathbf{E}_j(\mathbf{x}^{\text{dis}}_{t,j}) + \sum\limits_{j = 1}^{C_{\text{con}}} \psi_j(\mathbf{x}^{\text{con}}_{t,j}), & t \in [L + 1 : L + H] \quad \text{ (Forecast Window)}
    \end{cases}
\end{equation}
%This additive operation integrates information across heterogeneous variables without introducing additional parameters. 
During the forecast window, where $\mathbf{y}_t$ is the target to be predicted, only exogenous variables are used to construct $\mathbf{Z}_t$. The embedding $\mathbf{Z}_{1:L+H} = \{\mathbf{Z}_1,\cdots,\mathbf{Z}_t, \cdots, \mathbf{Z}_{L+H}\} \in \mathbb{R}^{(L+H)\times d}$ 
now encodes a rich amount of information from endogenous and exogenous variables. However, such a rich embedding may also introduce noise, as irrelevant information from heterogeneous inputs can be mixed into the representations. To mitigate this, ProtoTS introduces a bottleneck layer to filter out irrelevant information while retaining the most predictive components. Concretely, we project the high-dimensional embeddings into a lower-dimensional latent space through a bottleneck projection, $\mathbb{R}^{d} \to \mathbb{R}^{d_{\text{bottle}}} \to \mathbb{R}^{d}$, where $d_{\text{bottle}} \ll d$. The compressed representation is then mapped back to the original space, retaining the most relevant information while filtering out noise~\citep{wang2021variational}. We apply the bottleneck layer within an MLP-Mixer architecture~\citep{tolstikhin2021mlp}, inserting it into both $\text{MLP}_{\text{feature}}$ and $\text{MLP}_{\text{time}}$. Fusion is performed sequentially, first along the feature dimension and then along the temporal dimension, $\text{T}$ denotes the transpose of the vector:
\begin{equation}
\mathbf{Z}_{1:L+H}^{(l+1)} = \text{MLP}_{\text{time}}(\text{MLP}_{\text{feature}}(\mathbf{Z}_{1:L+H}^{(l)}) ^ \text{T}) ^ \text{T}.
\end{equation}
The bottleneck fusion layer is stackable, where $l$ denotes the layer index. After stacking, we obtain $\mathbf{Z}_{1:L+H} \in \mathbb{R}^{(L+H) \times d}$ in the original embedding dimension. To enable prototype matching in a unified one-dimensional vector space, we perform a linear aggregation along the temporal dimensions:
\begin{equation}
\hat{\mathbf{Z}} = \mathbf{Z}_{1:L+H}^{\text{T}} \text{W}, \hat{\mathbf{Z}} \in \mathbb{R}^{d}
\end{equation}
where $\text{W} \in \mathbb{R}^{(L+H)\times 1}$ projects the sequence length from $L+H$ to 1.

\textbf{Prototype Similarity Computing}\quad To provide a user-friendly global explanation while handling complex information at a local level for detailed interpretation and accurate forecasting, ProtoTS integrates prototypical explanation directly into its prediction mechanism via a learnable set of prototypes $\mathbf{\Pi}$~\citep{zhang2024prototypical}. Each prototype in $\mathbf{\Pi}$ is represented as an embedding $\boldsymbol{\mu} \in \mathbb{R}^{d}$ and its corresponding temporal pattern $\mathbf{p} \in \mathbb{R}^{T}$, which are both learnable parameters. Here, $\boldsymbol{\mu}$ is used to compute the similarity between the query representation $\hat{\mathbf{Z}}$ and the prototypes, by applying a softmax over their distances $d(\hat{\mathbf{Z}}, \boldsymbol{\mu})$. The distance can be chosen flexibly, such as Euclidean distance, cosine similarity. We adopt Euclidean distance, under which the prototypes are equivalent to a Gaussian mixture model in the embedding with an identity covariance matrix~\citep{allen2019infinite}. Meanwhile, temporal pattern $\mathbf{p}$ is used to form the final prediction. Specifically, we fix each prototype temporal pattern $\mathbf{p}$ over a predefined period of length $T$, and during forecasting, the prototype pattern is aligned with the forecast window $H$ based on its phase and length~\citep{lin2024cyclenet}.

\subsection{Hierarchical Prototype Learning}
\label{sec:hierarchical}

The key challenge in designing the prototype set $\mathbf{\Pi}$ lies in balancing predictive performance with interpretability. Using too few prototypes may lead to poor forecasting accuracy, while employing too many can compromise interpretability by introducing overly similar patterns that are hard to distinguish and reason about. To address this trade-off, ProtoTS adopts a hierarchical prototype learning strategy that organizes the prototype set in a tree structure, effectively combining modeling capacity with multi-level interpretability. 

\textbf{Root Level Learning}\quad At the root level, ProtoTS aims to capture the overall pattern using only a small set of prototypes, thus ensuring interpretability by allowing users to efficiently form a global understanding. These root prototypes typically correspond to coarse-grained temporal patterns that consistently recur over long horizons (e.g., seasonal trends, holiday effects; see Figure~\ref{casestudy}(Layer-1)). The similarity of root prototype $c$ is computed based on $\boldsymbol{\mu}_{c}$, followed by a softmax operation:
\begin{equation}
f(\hat{\mathbf{Z}}| \boldsymbol{\mu}_c) = \frac{\exp(-d(\hat{\mathbf{Z}}, \boldsymbol{\mu}_c))}{\sum_{i=1}^{N} \exp(-d(\hat{\mathbf{Z}}, \boldsymbol{\mu}_i))}.
\end{equation}
The final prediction $\widehat{\mathbf{Y}}$ is then generated as a weighted combination of the temporal patterns from all root prototypes:
\begin{equation}
\widehat{\mathbf{Y}}_{L+1:L+H} = \sum_{i=1}^{N} f(\hat{\mathbf{Z}}| \boldsymbol{\mu}_i) \cdot \mathbf{p}_i .
\end{equation}
At the first stage, the model is trained using the root prototypes until convergence. This root-level learning allows the model to establish a globally interpretable set of representative temporal patterns, providing users with an intuitive understanding of the model’s decision logic at a high level. 

\textbf{Splitting Strategy}\quad Once the model converges, each root prototype is further split into $M$ child prototypes to introduce fine-grained variations. For root prototype $i$, its $M$ child prototypes are denoted as $\{(\boldsymbol\mu_{i,j}, \mathbf{p}_{i,j})\}_{j=1}^M$. The similarity between $\hat{\mathbf{Z}}$ and its child prototypes is computed within this local group, ensuring all child prototypes remain associated with their root:
\begin{equation}
f(\hat{\mathbf{Z}}| \boldsymbol\mu_{i,k}) = \frac{\exp(-d(\hat{\mathbf{Z}}, \boldsymbol\mu_{i,k}))}{\sum_{j=1}^{M} \exp(-d(\hat{\mathbf{Z}}, \boldsymbol\mu_{i, j}))}. 
\end{equation}
Based on this hierarchical matching, the final prediction is formed as a weighted combination of all child prototypes, modulated by both root-level and child-level similarities. The complete forecasting output is given by:
\begin{equation}
\label{2levelpredict}
\widehat{\mathbf{Y}}_{L+1:L+H} = \sum_{i=1}^{N} f(\hat{\mathbf{Z}}|\boldsymbol\mu_i)  \sum_{j=1}^{M} f(\hat{\mathbf{Z}}|\boldsymbol\mu_{i, j})\cdot\mathbf{p}_{i, j}. 
\end{equation}

\begin{algorithm}[H]
\caption{Prototype Splitting Rule}
\label{splitting-rule}
\hrule
\vspace{4pt}

\textbf{Input:} Training dataset $\mathcal{D}$, leaf prototypes $\Pi_{\text{leaf}}=\{\text{leaf}_1,\dots \text{leaf}_{n_{\text{leaf}}}\}$, 
top-$k$ activation count, selection ratio $\alpha$.
\par % ←—————— 强制换行

\textbf{Initialize:} $\text{Loss}[l] \gets 0$, $\text{Count}[l] \gets 0$ for all leaf prototypes $\text{leaf}_l$.
\par

\textbf{For each} training instance $(x, y)$ in $\mathcal{D}$:
\par
\quad Compute prediction $\hat{y}$ and instance loss $L = \mathrm{MAE}(y,\hat{y})$.
\par
\quad Compute similarity scores $\{s_l\}_{l=1}^{n_{\text{leaf}}}$ over all leaf prototypes.
\par
\quad Let $\mathcal{T}$ be indices of top-$k$ similar prototypes based on $s_l$.
\par
\quad \textbf{For each} $l \in \mathcal{T}$:
\par
\qquad $\text{Loss}[l] \gets \text{Loss}[l] + L$.
\par
\qquad $\text{Count}[l] \gets \text{Count}[l] + 1$.
\par

\vspace{6pt}
\textbf{Normalized loss:}
\[
\text{NormLoss}[l] =
\begin{cases}
\frac{\text{Loss}[l]}{\text{Count}[l]}, & \text{if }\text{Count}[l] > 0, \\
0, & \text{otherwise}.
\end{cases}
\]

Select leaf prototypes with top $\alpha\%$ values in $\text{NormLoss}[l]$.
\vspace{2pt}
\hrule
\end{algorithm}

While the above describes a two-level hierarchy, the structure is general: any leaf prototype can be incrementally split to form a multi-level tree. However, not all prototypes need splitting, and some splits may yield overly similar temporal patterns. Therefore, we propose a splitting rule for determining which leaf prototypes should be further refined. The rule is designed to identify prototypes whose current temporal patterns are insufficiently representative of their associated instances and thus require additional fine-grained refinement. The pseudocode is provided in Algorithm~\ref{splitting-rule}.

This hierarchical design enables experts to first obtain a global understanding through high-level root prototypes (e.g., distinguishing between seasonal or holiday-related patterns), and then refine their insights by examining lower-level child prototypes that capture more localized variations (e.g., differentiating between long holidays and short holidays). Moreover, the hierarchy empowers expert interaction. Experts can steer the model by selectively splitting specific prototypes, introducing new root-level prototypes, or directly editing the temporal pattern to integrate domain knowledge. 

\textbf{Loss Function}\quad In addition to the standard L1 forecasting loss, we incorporate an entropy-based regularization on the prototype weights $f(\hat{\mathbf{Z}}\mid\boldsymbol{\mu}_i)$, with $\lambda$ controlling the strength, to encourage a few main prototypes to cover most predictions. We note that an L2 loss can also be used.
\begin{equation}
\label{loss_function}
\mathcal{L} =||\widehat{\mathbf{Y}}_{L+1:L+H}-{\mathbf{Y}}_{L+1:L+H}||_{1}-\lambda\sum_{i=1}^{N} f(\hat{\mathbf{Z}}|\boldsymbol\mu_{i}) log(f(\hat{\mathbf{Z}}|\boldsymbol\mu_{i})). 
\end{equation}

\vspace{1em}
\section{Experiments}
We conduct extensive experiments to thoroughly evaluate the effectiveness of ProtoTS. The main results show that ProtoTS consistently delivers strong performance on time series forecasting with exogenous variables. ProtoTS also achieves high accuracy on the multivariate time series forecasting, as shown in Appendix~\ref{multi}. We further analyze the effectiveness of model components and the quantitative evaluation confirms that ProtoTS provides the most understandable and usable explanation. Finally, a case study illustrating our ability to provide global explanations and fine-grained understanding, while supporting expert editing to enhance both performance and interpretability.
% To demonstrate ProtoTS's effectiveness,  we run extensive experiments on time series forecasting with exogenous variables. These experiments confirm its predictive performance and share a case study demonstrating the interpretability and editability of ProtoTS. We also open-source LOF (\textbf{LO}ad \textbf{F}orcasting dataset), which contains abundant covariates and a real application scenario, to inject fresh energy into the field.

\subsection{Overall Performance}

\textbf{Datasets}\quad Load Forecasting dataset (LOF) is a synthetic electric load dataset, including 22 supporting covariates and covering four regions. We also evaluated on the Electricity Price Forecasting (EPF) dataset~\citep{lago2021forecasting}, which covers hourly electric prices from five Nordic markets. The specific details of the datasets can be found in the Appendix~\ref{dataset-descriptions}.

\textbf{Baselines}\quad To provide a wide‑ranging comparison, we benchmark eight models in total: five state-of-the-art deep-learning methods (TimeXer~\citep{wang2024timexer}, iTransformer~\citep{liu2023itransformer}, TiDE~\citep{das2023long}, TFT~\citep{lim2021temporal}, N‑BEATSx~\citep{olivares2023neural}) together with three strong machine-learning baselines (XGBoost~\citep{chen2016xgboost}, LightGBM~\citep{ke2017lightgbm}, pyGAM~\citep{serven2018pygam}). All deep-learning models support exogenous variables as inputs, and we further extend TimeXer and iTransformer to incorporate future covariates. The machine learning models generate forecasts recursively with exogenous inputs.

% Full results for time series forecasting with exogenous variables are presented in Appendix~\ref{full-results}. Additionally, we evaluate the performance of our method on multivariate time series forecasting, with the corresponding results provided in Appendix~\ref{multi}.

\textbf{Results}\quad The MAE on the LOF dataset is shown in Table~\ref{LOFresults-MAE}, ProtoTS achieves state-of-the-art performance across four datasets, reducing MSE by 48.3\% and MAE by 20.9\% over the best baseline. Interpretable models like N-BEATSx are sensitive to outliers, resulting in high MSE. Transformer-based models fail to capture heterogeneous exogenous information due to single-channel encoding. Although TiDE incorporates exogenous variables in both encoder and decoder, its performance is hindered by irrelevant noise. Machine learning models perform well on stable datasets but struggle with complex scenarios. In contrast, ProtoTS effectively captures diverse information and filters noise, achieving superior performance. Visual comparison can be found in Appendix~\ref{Comparison}. 

% discrimination power

We also evaluate ProtoTS on the EPF dataset. Unlike LOF, this task involves fewer covariates but exhibits higher volatility, posing challenges for accurate forecasting. As shown in Table~\ref{EPFresults-MAE}, ProtoTS achieves the best performance across all five markets, reducing MSE by 8\% and MAE by 8\% compared to the best-performing baseline. We further conduct a sensitivity analysis under 5 random seeds, see Appendix~\ref{error-analysis}. Full implementation details can be found in the Appendix~\ref{implementation-details}.

\newpage

\begin{table}[H]
\caption{MAE results on the LOF dataset with 22 supporting covariates. The look-back window and forecast window are set to 384 and 96 for all methods. $\Delta$ means the relative improvement of ProtoTS over other baselines. Full MSE results can be found in Appendix~\ref{full-results-LOF}.}
\label{LOFresults-MAE}
\centering
\scalebox{0.76}{
\begin{tabular}{c|c|c|c|c|c|c|c|c|c|c}

\toprule

\multicolumn{2}{c|}{Model} & \hspace{0.5mm}ProtoTS\hspace{0.5mm} & \hspace{0.2mm}TimeXer\hspace{0.2mm}& \hspace{2mm}iTrans.\hspace{2mm} & \hspace{3mm}TiDE\hspace{3mm} & \hspace{3.1mm}TFT\hspace{3.1mm} & NBEATSx & XGBoost & LightGBM & \hspace{0.1mm}pyGAM\hspace{0.1mm} \\

\midrule

\multirow{4}{*}[-12pt]{LOF} & RE & \textbf{0.198} &  0.272 &  0.279 &  0.253 &  0.342 &  0.388  &  0.405 &  0.366 &  0.388  \\

\cmidrule{2-11} 
& YC &\textbf{0.055}& 0.079& 0.080&0.057&0.116&0.572 &0.084&0.082&0.117\\

\cmidrule{2-11} 

& EA &\textbf{0.059}&0.096 &0.097&0.061 &0.108 &0.589 &0.092&0.085&0.108 \\

\cmidrule{2-11} 

& PC &\textbf{0.112} &0.182 &0.139 & 0.164 & 0.285& 0.704 & 0.230 &0.222 &0.269 \\

\midrule

\multicolumn{2}{c|}{Avg} & \textbf{0.106}  & 0.157 & 0.149  & 0.134&0.213 &0.563 & 0.203 &0.189 &0.221 \\

\midrule

\multicolumn{2}{c|}{$\Delta$} & -  & 32\% & 29\%  & 21\% & 50\%  & 81\% & 48\%  & 44\%  & 52\% \\

\bottomrule
\end{tabular} }
\end{table}

% , which focuses on electricity price forecasting with six exogenous variables. 
% These results highlight ProtoTS ability to effectively model exogenous influences despite limited covariates, further demonstrating its robustness and generalizability.

\vspace{-1.5em}

\begin{table}[H]
\caption{MAE results on EPF dataset equipped with 6 supporting covariates. The look-back window and forecast window are set to 168 and 24 for all methods. $\Delta$ means the relative improvement of ProtoTS over other baselines. Full MSE results can be found in Appendix~\ref{full-results-EPF}}
\label{EPFresults-MAE}
\centering
\scalebox{0.765}{
\begin{tabular}{c|c|c|c|c|c|c|c|c|c|c}

\toprule

\multicolumn{2}{c|}{Model} & \hspace{0.5mm}ProtoTS\hspace{0.5mm} & \hspace{0.2mm}TimeXer\hspace{0.2mm}& \hspace{2mm}iTrans.\hspace{2mm} & \hspace{3mm}TiDE\hspace{3mm} & \hspace{3.1mm}TFT\hspace{3.1mm} & NBEATSx & XGBoost & LightGBM & \hspace{0.1mm}pyGAM\hspace{0.1mm} \\

\midrule

\multirow{5}{*}[-12pt]{EPF} & NP  & \textbf{0.213}&0.240 & 0.264 & 0.318 & 0.277  &0.266  & 0.488 & 0.431 & 0.480 \\

\cmidrule{2-11} 

& PJM &\textbf{0.152}& 0.173 & 0.171&0.226 &0.226 &0.182  & 0.292  & 0.256 & 0.351 \\

\cmidrule{2-11}

& BE  &\textbf{0.226}&0.241 &0.266  &0.323 & 0.277& 0.324 & 0.451  & 0.392  & 0.607 \\

\cmidrule{2-11}

& FR &\textbf{0.183}& 0.192&0.213 &0.281 & 0.242&0.340 & 0.374 & 0.361 & 0.570 \\

\cmidrule{2-11}

& DE &\textbf{0.318}& 0.343 & 0.335 & 0.474&0.489 & 0.408 & 0.539  & 0.506 & 0.551 \\

\midrule

\multicolumn{2}{c|}{Avg}  & \textbf{0.218} & 0.238 & 0.250 & 0.324 & 0.302 & 0.304  & 0.429 & 0.389 & 0.512 \\

\midrule

\multicolumn{2}{c|}{$\Delta$} & - & 8\% & 13\% & 33\% & 28\% & 28\% & 49\% & 44\% & 57\% \\

% \midrule

% \multicolumn{2}{c|}{EDF} & & & & & & & & & & & & & & & & \\

\bottomrule

\end{tabular} }
\end{table}

\subsection{Ablation Study and Sensitivity Analysis}
\textbf{Ablation Study}\quad We conduct an ablation study to assess the contribution of the three key components in ProtoTS: multi-channel embedding, bottleneck layer and hierarchical structure. To assess the modeling of heterogeneous variables, we perform ablations by simplifying the embedding to a single channel, removing the bottleneck layer, and flattening the hierarchical prototype structure. The ablation results in Table~\ref{ablationstudy} demonstrate the importance of each key component in ProtoTS. While removing any component degrades performance, the full model with all three achieves the best results, confirming the effectiveness of the overall design.

\begin{table}[h]
\centering
\caption{Ablation Results. Evaluating the performance after removing one of the following components: bottleneck layer, multi-channel embedding, and hierarchical structure.}
\label{ablationstudy}
\scalebox{0.85}{
\begin{tabular}{c|cc|cc|cc|cc|cc}
\toprule
Models & \multicolumn{2}{c|}{PC} & \multicolumn{2}{c|}{YC} & \multicolumn{2}{c|}{RE} & \multicolumn{2}{c|}{EA} & \multicolumn{2}{c}{AVG} \\
\cmidrule{2-11}
Metric & MSE & MAE & MSE & MAE & MSE & MAE & MSE & MAE & MSE & MAE \\
\midrule
w/o bottleneck & 0.044 & 0.160 & 0.013 & 0.089 & 0.129 & 0.248 & 0.010 & 0.073 & 0.049 & 0.143 \\
 w/o multi-channel  & 0.034 & 0.130 & \textbf{0.006} & 0.058 & 0.108 & 0.216 & \textbf{0.007} & 0.062 & 0.039 & 0.117\\
 w/o hierachy & 0.026 & 0.117 & \textbf{0.006} & 0.060 & 0.089 & 0.202 & \textbf{0.007} & 0.061 & 0.032 & 0.110 \\
ProtoTS & \textbf{0.025} & \textbf{0.112} & \textbf{0.006} & \textbf{0.055} & \textbf{0.085} & \textbf{0.198} & \textbf{0.007} & \textbf{0.059} & \textbf{0.031} & \textbf{0.106} \\
\bottomrule
\end{tabular}}
\end{table}

\begin{figure}[H]
    \centering
    \vspace{-1em}
    \begin{adjustbox}{center,raise=0cm,margin*=-0.2cm 0cm 0cm 0cm}
        \includegraphics[width=1\linewidth]{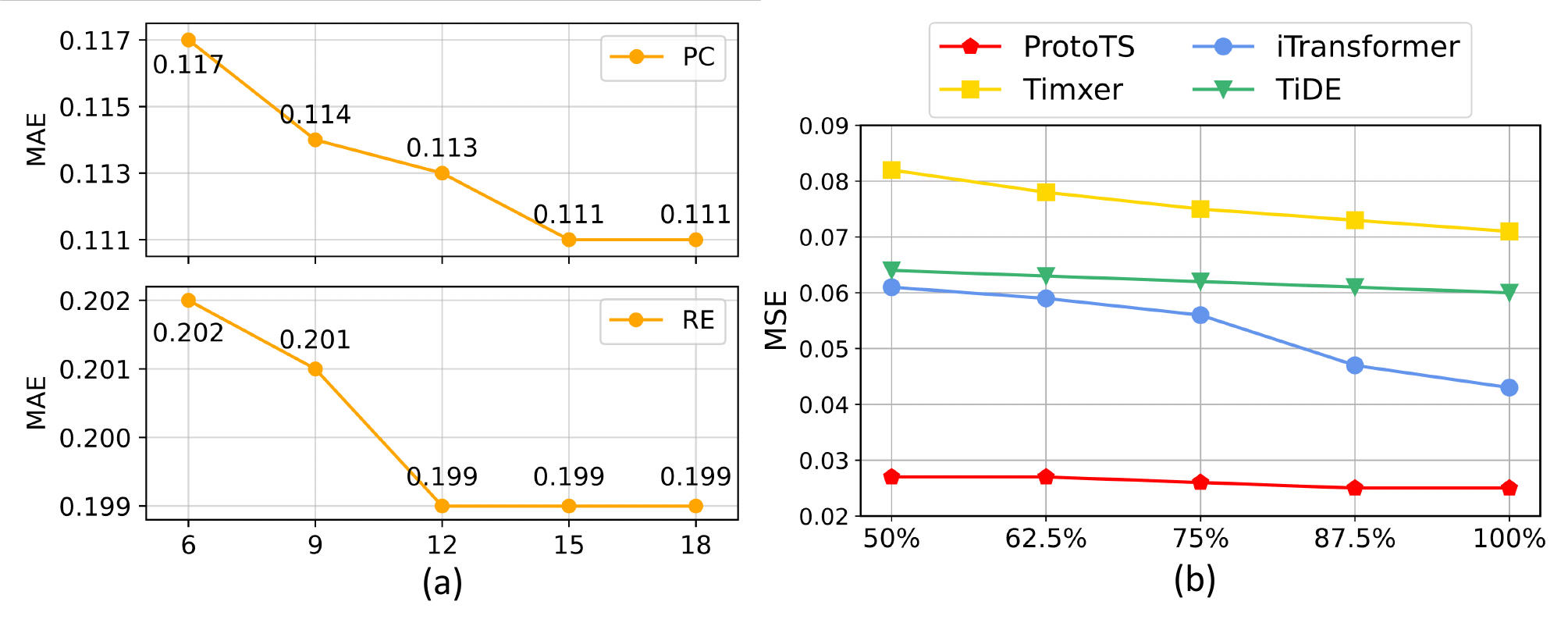}
    \end{adjustbox}
    \vspace{-1.6em}
    \caption{Sensitivity analysis: (a) Effect of increasing root prototype numbers from \{6, 9, 12, 15, 18\}. (b) Data efficiency as the training data proportion increases from 50\% to 100\%.}
    \vspace{-1em}
    \label{modelanalysis}
\end{figure}

\textbf{Number of Root Prototypes}\quad We evaluate how the number of root prototypes affects model performance, as shown in Figure~\ref{modelanalysis}(a). The results clearly show that prototypes number is correlated with forecasting accuracy. Increasing the number of prototypes leads to lower MAE, as more prototypes enable the model to capture a richer variety of temporal patterns for better representation. However, once the number of root prototypes reaches a threshold, the performance no longer improves. This indicates that the prototypes have already covered all typical patterns in the dataset.

\textbf{Proportion of Train Data}\quad To evaluate the impact of limited training data on model performance, we gradually increase the size of the PC dataset from 50\% to 100\%. As the available data decreased, existing baselines such as TimeXer, iTransformer, and TiDE show clear drops in MSE. In contrast, ProtoTS maintains stable results with only minor performance loss, which improves high data efficiency~\citep{jin2023prototypical}. The comparison results are illustrated in Figure~\ref{modelanalysis}(b).

\begin{wraptable}{r}{0.45\linewidth}
\centering
\vspace{-2em}
\caption{Quantitative Evaluation.}
\label{Quantitative-Evaluation}
\begin{tabular}{ccc}
\toprule
\textbf{Model} & \textbf{UPrec} $\uparrow$ & \textbf{SUS Score} $\uparrow$ \\
\midrule
ProtoTS & \textbf{77 $\pm$ 3.6\%} & \textbf{73.36} \\
TFT & 64 $\pm$ 3.4\% & 29.74 \\
NBEATSx & 62 $\pm$ 2.8\% & 38.66 \\
\bottomrule
\end{tabular}
\vspace{-1em}
\end{wraptable}

\subsection{Quantitative Evaluation of Interpretability}
\label{user-study}
In this section, we conduct a quantitative evaluation of interpretability. We aim to answer the two questions: 1) How understandable and accurate are the prototypes in explaining the predictions? 2) How usable are the explanation systems provided by different interpretable models (ProtoTS, TFT, NBEATSx)? Specifically, we provide 24 users with different explanations as well as the corresponding predictions, where the three models are presented randomly to avoid bias. Users were asked concrete questions about the input variables, such as "Given this prediction and explanations, which season does the predicted day most likely belong to?". To evaluate interpretability, we define User Precision (UPrec) as the proportion of correctly answered variable-related questions, reflecting whether users can accurately infer the values of relevant input features from the explanations~\citep{lee2025toward}. To further assess usability, we adopt the System Usability Scale (SUS)~\citep{brooke1996sus}, a widely used questionnaire that measures users’ perceived ease of use and overall satisfaction with a system. The quantitative results are reported in Table~\ref{Quantitative-Evaluation}, with additional details in Appendix~\ref{user-study-details}. 

\subsection{Case Study of Interpretability and Steerability
%: Interpretability Analysis on the QD Dataset
}
\label{sec-casestudy}
\begin{figure}[t]
    \centering
    \begin{adjustbox}{center,raise=0cm,margin*=-0.4cm 0cm 0cm 0cm}
        \includegraphics[width=1.03\linewidth]{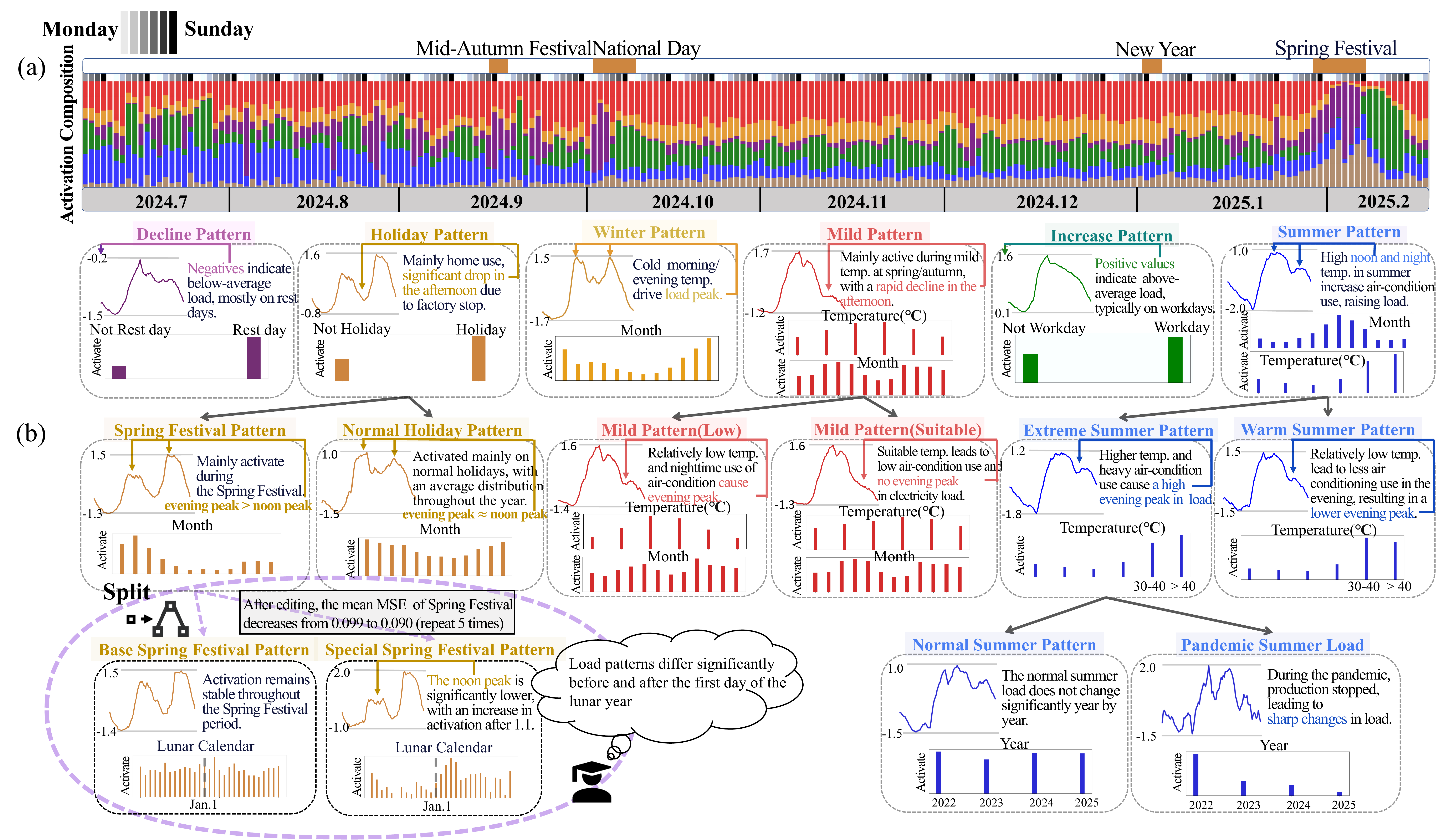}
    \end{adjustbox}
    \vspace{-1em}
    \caption{Case study of interpretability and steerability: (a) Activation patterns of prototypes, which each colored bar shows the degree of activation of its corresponding prototype each day. This visualization shows seasonal, weekly, and holiday-related prototype activations. (b) The learned prototype hierarchy that illustrates typical temporal patterns and their correlations with covariates for interpretable load forecasting. After the expert splits the first prototype at layer-2, the mean MSE across five random seeds decreases from 0.099 to 0.090 for Spring Festival related predictions.}
    \vspace{-1em}
    \label{casestudy}
\end{figure}

Figure~\ref{casestudy} illustrates how ProtoTS discovers meaningful and interpretable hierarchical prototypes that help better understand electricity load forecasting and facilitate expert edits that improves model performance. We use the RE case in LOF. The load is shaped by multiple external factors, including \textit{seasonal}, \textit{temperature} and \textit{weekday/holiday}. ProtoTS is initialized with 6 root prototypes, each further refined into 2 child prototypes. This hierarchy captures both coarse and fine temporal patterns. Time series predictions are weighted combinations of these prototypes (Figure~\ref{casestudy}(a)). The activation weights link each prototype's contribution to predictions and are also the key to interpretation.

\textbf{Obtaining overall understanding of temporal patterns}. The root prototypes provide coarse temporal patterns of high-level concepts (Figure~\ref{casestudy}(b), layer-1): workdays, non-workdays, holidays, winter, mild temperature, and summer. For example, the summer pattern (blue) exhibits high daytime peaks due to air-conditioning dominance. Autumn and winter data are dominated by distinct seasonal load profiles in red and orange. The brown prototype with a dual-peak pattern is active during holidays. This contrasts with the green weekday pattern with a single afternoon peak. The dual peaks suggest that industrial activity is paused, and the load is driven by residential behavior. 

\textbf{Fine-grained analysis of detailed temporal patterns.} The child prototypes provide fine-grained refinements of these general patterns (Figure~\ref{casestudy}(b), layer-2). For example, the root holiday prototype is split into two children: one capturing the pattern of the major holiday, Spring Festival (Chinese New Year), and the other representing the rest of the normal holidays. The Spring Festival prototype shows a more pronounced evening peak and flattened noon peak. 

\textbf{Steering the model to improve accuracy}. Editing the model for improvements becomes feasible thanks to the clearly interpretable temporal patterns learned in protoTS (Figure~\ref{casestudy}(b), layer-3). Based on domain knowledge that the Spring Festival itself contains distinct sub-patterns of pre- and mid-holiday behaviors, we manually split the Spring Festival prototype into two prototypes, which learns a base pattern and a special pattern. The base pattern stays consistently active through the whole Spring Festival period. The special pattern starts from the Lunar Jan. $1^{\text{st}}$, strengthening progressively day by day, reflecting a deepening holiday atmosphere: reduced daytime electrical usage, and stronger evening peaks as people stay home. This edit reduced the MSE during Spring Festival by 0.009, showing how expert adjustments can enhance both accuracy and explainability. \looseness=-1

\vspace{-3mm}
\section{Conclusion}
\vspace{-3mm}

In this work, we proposed ProtoTS, a novel forecasting framework that achieves both high accuracy and multi-level interpretability through hierarchical prototypes. It learns low-noise representations from heterogeneous variables and matches them with prototypes to generate predictions and explanations, while the hierarchy enables coarse-to-fine pattern learning and expert-steerability. Extensive experiments and case studies on realistic datasets show state-of-the-art performance and transparent support, enabling experts to refine model behavior and bridging accuracy with interpretability.

% ProtoTS learns a low-noise overall representation enriched with relevant information from heterogeneous variables, and matches it with learned prototypes to generate predictions and provide detailed local interpretation. The prototype set offers a global understanding of typical temporal patterns present in the dataset, enabling users to grasp the model's decision logic at a higher level. Furthermore, prototypes are organized hierarchically, allowing coarse-to-fine pattern learning that balances predictive performance with expert-steerable global and local interpretability. Extensive experiments on realistic datasets demonstrate that ProtoTS consistently outperforms existing baselines, achieving state-of-the-art results while providing transparent and actionable decision-making support. Case studies further show how ProtoTS empowers domain experts to understand and refine model behavior through prototype-level interactions, bridging the gap between forecasting accuracy and interpretability in complex time series scenarios.

\newpage

\section*{Acknowledgements}
This work was partly supported by the National Natural Science Foundation of China (NSFC) (NO. 62476279, NO. U2436209), Major Innovation \& Planning Interdisciplinary Platform for the “Double-First Class” Initiative, Renmin University of China, the Fundamental Research Funds for the Central Universities, and the Research Funds of Renmin University of China No.24XNKJ18. This work was partially supported by fund for building world-class universities (disciplines) of Renmin University of China and Public
Computing Cloud, Renmin University of China. This work was partially supported by DAMO Academy (Hupan Laboratory) through DAMO Academy (Hupan Laboratory) Innovative Research Program.
% \section*{Ethics statement}

% We provide an ethics statement as follows:
% \begin{itemize}[nosep,leftmargin=1em,labelwidth=*,align=left]
%     \item Human subjects research: Our experiments include a user study (see Section~\ref{user-study}). The study does not raise any ethical concerns, and all participants gave informed consent prior to participation.
%     \item Datasets: All datasets used in our work are publicly available. The released dataset is a simulation of real-world data and does not contain any actual sensitive information.
%     \item Fairness and bias: Our work does not raise fairness or bias concerns.
% \end{itemize}

% \section*{Reproducibility statement} 

% We confirm that the contents of the main text and the appendix are sufficient to reproduce our work and our experiments are reproducible. Section~\ref{method} clearly describes the proposed method and workflow. Appendix~\ref{dataset-descriptions} provides detailed dataset descriptions, and Appendix~\ref{implementation-details} provides sufficient implementation details (including hyperparameters and training settings). We also include in the abstract a link to an anonymous repository containing our source code and the released dataset; the repository provides scripts for data preparation, training, and evaluation, along with configuration files and random seeds. We have verified that the code is consistent with the descriptions in the paper and that the reported experiments can be reproduced following the provided instructions.

\bibliography{iclr2026_conference}
\bibliographystyle{iclr2026_conference}

\newpage

\appendix

\section{Use of LLMs}
We used large language models (LLMs) only to aid or polish writing. Specifically, LLMs were employed to: (1) check grammar and spelling; (2) assist with professional academic English phrasing without hurting the original meaning; and (3) perform light-length reduction to meet formatting constraints. LLMs were not used to generate ideas, methods, experiments, figures, or results. LLMs were not used to find related works or to retrieve references. All suggestions were reviewed and verified by the authors, who take full responsibility for the final text.

\section{Implementation Details}

\subsection{Dataset Descriptions}
\label{dataset-descriptions}

\textbf{LOF dataset}\quad  The LOF dataset is an synthetic electric load forecasting dataset. It is designed to predict future electricity demand at the provincial and city levels, enable electricity generators, distributors, and suppliers to plan effectively ahead and promote energy conservation among the users~\citep{nti2020electricity}. Due to the influence of human industrial and commercial activities, electricity demand exhibits a clear daily cycle, maintaining a relatively stable and predictable pattern under normal conditions. The difficulty of load forecasting mainly arises from special instances that are strongly correlated with a large amount of exogenous information, such as time-related attributes and comprehensive weather data. Consequently, accurate electricity load forecasting requires models to not only effectively handle these variables but also provide insights into how they affect prediction results. Hence, electricity load forecasting is a typical task of time series forecasting with exogenous variables, characterized by a high demand for interoperability. The LOF dataset ranges across three years, with a sampling interval of 15 minutes. The LOF dataset includes 22 exogenous variables, consisting of 14 discrete time-related variables and 8 continuous weather variables. The specific details of these exogenous variables are presented in Table~\ref{LOF-covariates}.

\textbf{EPF dataset}\quad The EPF dataset is an electricity price forecasting datasets~\citep{lago2021forecasting}. It collects from five distinct day-ahead electricity markets in Northern Europe, covering a six year period with a sampling interval of 1 hour. The five datasets are described as follows: (1) NP corresponds to the Nord Pool market, providing hourly electricity prices, grid load data, and wind power forecasts from 2013-01-01 to 2018-10-24. (2) PJM refers to the Pennsylvania-New Jersey-Maryland market, including zonal electricity prices for the Commonwealth Edison (COMED) zone, along with system-wide load and COMED-specific load forecasts, covering the period from 2013-01-01 to 2018-10-24. (3) BE represents the Belgian electricity market, offering hourly price data, load forecasts for Belgium, and generation forecasts for France, collected from 2011-01-09 to  2016-10-31. (4) FR pertains to the French electricity market, comprising hourly price records, load forecasts, and generation forecasts over the timeframe from 2012-01-09 to 2017-10-31. (5) DE covers the German electricity market, documenting hourly electricity prices, zonal load forecasts for the TSO Amprion area, as well as wind and solar generation forecasts, with data available from 2012-01-09 to 2017-10-31.

\subsection{Implementation Details}
\label{implementation-details}
All experiments are implemented using PyTorch~\citep{paszke2019pytorch} and executed on a single NVIDIA V100 GPU with 32GB of memory. For model optimization, we employ the Adam~\citep{kingma2014adam} optimizer with an initial learning rate of and use our Loss as the objective function. The training is set for a maximum of 30 epochs, with early stopping applied to prevent overfitting. In our proposed model, the number of bottleneck fusion blocks is selected from the set \{1, 2, 3\}, while the dimension of the series representations, denoted as model dimensions, is chosen from \{32, 64, 128, 256, 512\}, the number of initial prototypes is searched from \{6, 8, 12\}, the deepest level is chosen from \{1, 2, 3\}, the number of each prototype can split is selected from the set \{2, 3\}. All baseline models used for comparison are re-implemented based on the TimeXer Repository benchmark. All baseline use REVIN~\citep{kim2021reversible} in training and test process.

\section{Full Results}
\subsection{Full Results of LOF dataset}
\label{full-results-LOF}
Full results of the LOF dataset can be found in Table~\ref{LOFresults}, including both MSE and MAE metrics. ProtoTS achieves state-of-the-art performance across all four datasets, reducing MSE by 48.3\% and MAE by 20.9\%. TimeXer, iTransformer, and TiDE show the most competitive results, with TiDE performing the best among them. Machine learning methods achieve decent performance with very short training time.

\begin{table}[H]
\caption{Full results of LOF dataset on the time series forecasting with exogenous variables task, equipped with 22 supporting covariates. The look-back window and forecast window are set to 384 and 96 for all baselines. $\Delta$ means the relative improvement of ProtoTS over other baselines.}
\label{LOFresults}
\centering

\scalebox{0.83}{
\begin{tabular}{c|c|c c|c c|c c|c c|c c|c c}

\toprule

\multicolumn{2}{c|}{Model} & \multicolumn{2}{c|}{RE} & \multicolumn{2}{c|}{YC}& \multicolumn{2}{c|}{EA} & \multicolumn{2}{c|}{PC} & \multicolumn{2}{c|}{AVG} &  \multicolumn{2}{c}{$\Delta$}\\

\cmidrule{3-14}

\multicolumn{2}{c|}{Metric} & MSE & MAE  & MSE & MAE & MSE & MAE & MSE & MAE & MSE &MAE & MSE &MAE\\

\midrule

\multicolumn{2}{c|}{ProtoTS} & \textbf{0.085} &\textbf{0.198} & \textbf{0.006} & \textbf{0.055} & \textbf{0.007} & \textbf{0.059} & \textbf{0.025} & \textbf{0.112} & \textbf{0.031} & \textbf{0.106} &- &- \\

\midrule

\multicolumn{2}{c|}{TimeXer} & 0.167 & 0.272 & 0.011 & 0.079 & 0.015 & 0.096 & 0.073 & 0.182 & 0.067 & 0.157 &53\% & 32\% \\

\midrule

\multicolumn{2}{c|}{iTransformer} & 0.169 & 0.279 & 0.013 & 0.080 & 0.018 & 0.097 & 0.043 & 0.139 & 0.061 & 0.149 & 49\% & 29\% \\

\midrule
\multicolumn{2}{c|}{TiDE} & 0.158 & 0.253 & 0.007 & 0.057 & 0.008 & 0.061 & 0.066 & 0.164 & 0.060 & 0.134 & 48\% & 21\% \\

\midrule
\multicolumn{2}{c|}{TFT} & 0.228 & 0.342 & 0.025 & 0.116 & 0.021 & 0.108 & 0.154 & 0.285 & 0.107 & 0.213 & 71\% & 50\% \\

\midrule
\multicolumn{2}{c|}{NBEATSx} & 0.815 & 0.388 & 1.553 & 0.572 & 2.185 & 0.589 & 2.467 & 0.704 & 1.755 & 0.563 & 98\% & 81\% \\

\midrule
\multicolumn{2}{c|}{XGBoost} & 0.300 & 0.405 & 0.014 & 0.084 & 0.017 & 0.092 & 0.101 & 0.230 & 0.108 & 0.203 & 71\% & 48\% \\

\midrule
\multicolumn{2}{c|}{LightGBM} & 0.253 & 0.366 & 0.012 & 0.082 & 0.014 & 0.085 & 0.094 & 0.222 & 0.093 & 0.189 & 67\% & 44\% \\

\midrule
\multicolumn{2}{c|}{pyGAM} & 0.263 & 0.388 & 0.022 & 0.117 & 0.019 & 0.108 & 0.126 & 0.269 & 0.108 & 0.221 & 71\% & 52\% \\

\bottomrule

\end{tabular}}

\end{table}

\subsection{Full Results of EPF dataset}
\label{full-results-EPF}
Full results of the EPF dataset can be found in Table~\ref{EPFresults}, including both MSE and MAE metrics. ProtoTS achieves state-of-the-art performance across all five datasets, reducing MSE by 8\% and MAE by 8\%. Similarly, TimeXer, iTransformer, and TiDE show competitive results, with TimeXer performing the best in this case. Machine learning methods perform poorly on EPF, where the number of covariates is limited.

\begin{table}[H]
\caption{Full results of EPF dataset on the time series forecasting with exogenous variables task, equipped with 6 supporting covariates. The look-back window and forecast window are set to 168 and 24 for all baselines. $\Delta$ means the relative improvement of ProtoTS over other baseline.}
\label{EPFresults}

\scalebox{0.72}{
\begin{tabular}{c c|c c|c c|c c|c c|c c|c c|c c}

\toprule

\multicolumn{2}{c|}{Model} & \multicolumn{2}{c|}{NP} & \multicolumn{2}{c|}{PJM}& \multicolumn{2}{c|}{BE} & \multicolumn{2}{c|}{FR}& \multicolumn{2}{c|}{DE} & \multicolumn{2}{c|}{AVG} &  \multicolumn{2}{c}{$\Delta$}\\

\cmidrule{3-16} 

\multicolumn{2}{c|}{Metric} & MSE & MAE  & MSE & MAE & MSE & MAE & MSE & MAE & MSE &MAE & MSE &MAE& MSE & MAE\\

\midrule

\multicolumn{2}{c|}{ProtoTS} & \textbf{0.168} & \textbf{0.213} & \textbf{0.064} & \textbf{0.152} & \textbf{0.353} & \textbf{0.226} & \textbf{0.351} & \textbf{0.183} & \textbf{0.271} & \textbf{0.318} & \textbf{0.241} & \textbf{0.218} & - & - \\

\midrule

\multicolumn{2}{c|}{TimeXer} & 0.194 & 0.240 & 0.078 & 0.173 & 0.365 & 0.241 & 0.355 & 0.192 & 0.319 & 0.343 & 0.262 & 0.238 & 8\% & 8\% \\

\midrule

\multicolumn{2}{c|}{iTransformer} & 0.208 & 0.264 & 0.076 & 0.171 & 0.375 & 0.266 & 0.386 & 0.213 & 0.282 & 0.335& 0.265& 0.250& 9\% & 13\% \\

\midrule
\multicolumn{2}{c|}{TiDE} & 0.306 & 0.318 & 0.121 & 0.226 & 0.495 & 0.323 & 0.492 & 0.281 & 0.533 & 0.474 & 0.389& 0.324& 38\% & 33\% \\

\midrule
\multicolumn{2}{c|}{TFT} & 0.259 & 0.277 & 0.138 & 0.226 & 0.412 & 0.277 & 0.452 & 0.242 & 0.709 & 0.489 & 0.394 & 0.302 & 39\% & 28\% \\

\midrule
\multicolumn{2}{c|}{NBEATSx} & 0.222 & 0.266 & 0.079 & 0.182 & 0.507 & 0.324 & 1.920 & 0.340 & 0.430 & 0.408 & 0.632 & 0.304 & 62\% & 28\% \\

\midrule
\multicolumn{2}{c|}{XGBoost} & 1.002 & 0.488 & 0.276 & 0.292 & 0.817 & 0.451 & 0.622 & 0.374 & 0.618 & 0.539 & 0.669 & 0.429 & 64 \% & 49\% \\

\midrule
\multicolumn{2}{c|}{LightGBM} & 0.678 & 0.431 & 0.234 & 0.256 & 0.598 & 0.392 & 0.603 & 0.361 & 0.549 & 0.506 & 0.532 & 0.389 & 55 \% & 44\% \\

\midrule
\multicolumn{2}{c|}{pyGAM} & 0.599 & 0.480 & 0.281 & 0.351 & 1.028 & 0.607 & 1.116 & 0.570 & 0.671 & 0.551 & 0.739 & 0.512 & 67 \% & 57\% \\

\bottomrule

\end{tabular}}

\end{table}

\subsection{Sensitive Analysis}
\label{error-analysis}
To investigate the consistency of the model's performance across different randomized scenarios, we selected five random seeds and reported the mean and standard deviation of the model's performance. The experiments in Table~\ref{error-lof} and Table~\ref{error-epf} show that the model maintains stable performance without statistically significant variations in most cases.

\begin{table}[H]
\centering
\caption{Sensitive Analysis on LOF dataset. We report our performance in $\mathrm{mean}${\scriptsize $\pm$ $\mathrm{std}$} format.}
\label{error-lof}
\scalebox{0.75}{
\begin{tabular}{cc cc cc cc cc}
\toprule

\multicolumn{2}{c}{Model} & \multicolumn{2}{c}{ProtoTS} & \multicolumn{2}{c}{TimeXer} & \multicolumn{2}{c}{iTransformer} & \multicolumn{2}{c}{TiDE} \\

\cmidrule{3-10}

\multicolumn{2}{c}{Metric} & MSE & MAE & MSE & MAE & MSE & MAE & MSE & MAE \\

\midrule

\multicolumn{2}{c}{PC}& 0.025{\scriptsize $\pm$ 0.0011}& 0.113{\scriptsize $\pm$0.0032}&0.0754{\scriptsize $\pm$ 0.0035}&0.186{\scriptsize $\pm$0.0068}&0.043{\scriptsize $\pm$0.0006}&0.161{\scriptsize $\pm$0.0009}&0.066{\scriptsize $\pm$0.0001}&0.164{\scriptsize $\pm$0.0001} \\

\midrule
\multicolumn{2}{c}{YC}& 0.006{\scriptsize $\pm$ 0.0002}& 0.055{\scriptsize $\pm$ 0.0011}&0.012{\scriptsize $\pm$0.0016}&0.080{\scriptsize $\pm$0.0075}&0.014{\scriptsize $\pm$0.0017}&0.083{\scriptsize $\pm$0.0036}&0.007{\scriptsize $\pm$0.0001}&0.057{\scriptsize $\pm$0.0001} \\

\midrule
\multicolumn{2}{c}{RE}&0.093{\scriptsize $\pm$ 0.0074} &0.206{\scriptsize $\pm$ 0.0074}&0.179{\scriptsize $\pm$0.0141}&0.291{\scriptsize $\pm$0.0160}&0.169{\scriptsize $\pm$0.0094}&0.275{\scriptsize $\pm$0.0123}&0.158{\scriptsize $\pm$0.0001}&0.253{\scriptsize $\pm$0.0005} \\
\midrule
\multicolumn{2}{c}{EA}& 0.007{\scriptsize $\pm{}$0.0004}& 0.059{\scriptsize $\pm$0.0010}&0.015{\scriptsize $\pm$0.0022}&0.092{\scriptsize $\pm$0.0070}&0.018{\scriptsize $\pm$0.0024}&0.097{\scriptsize $\pm$0.0049}&0.008{\scriptsize $\pm$0.0001}&0.061{\scriptsize $\pm$0.0002} \\

\midrule
\multicolumn{2}{c}{AVG}& 0.033{\scriptsize $\pm$ 0.0022}& 0.108{\scriptsize $\pm$ 0.0031}&0.070{\scriptsize $\pm$ 0.00535}&0.162{\scriptsize $\pm$ 0.0093}&0.061{\scriptsize $\pm$ 0.0035}&0.154{\scriptsize $\pm$ 0.0054}&0.060{\scriptsize $\pm$ 0.0001}&0.134{\scriptsize $\pm$ 0.0002} \\

\bottomrule

\end{tabular}}
\end{table}

\begin{table}[H]
\centering
\caption{Sensitive Analysis on EPF dataset. We report our performance in $\mathrm{mean}${\scriptsize $\pm$ $\mathrm{std}$} format.}
\label{error-epf}
\scalebox{0.75}{
\begin{tabular}{cc cc cc cc cc cc}
\toprule

\multicolumn{2}{c}{Model} & \multicolumn{2}{c}{ProtoTS} & \multicolumn{2}{c}{TimeXer} & \multicolumn{2}{c}{iTransformer} & \multicolumn{2}{c}{TiDE} \\

\cmidrule{3-10}

\multicolumn{2}{c}{Metric} & MSE & MAE & MSE & MAE & MSE & MAE & MSE & MAE \\

\midrule

\multicolumn{2}{c}{NP}& 0.169{\scriptsize $\pm$ 0.0019} & 0.211{\scriptsize $\pm$ 0.0015}&0.194{\scriptsize $\pm$0.0036}&0.238{\scriptsize $\pm$0.0023}&0.212{\scriptsize $\pm$0.0029}&0.269{\scriptsize $\pm$0.0026}&0.316{\scriptsize $\pm$0.0081}&0.326{\scriptsize $\pm$0.0058} \\

\midrule

\multicolumn{2}{c}{PJM}& 0.067{\scriptsize $\pm$ 0.0013}& 0.153{\scriptsize $\pm$ 0.0020}&0.078{\scriptsize $\pm$ 0.0011}&0.171{\scriptsize $\pm$0.0012}&0.081{\scriptsize $\pm$ 0.0040}&0.171{\scriptsize $\pm$0.0014}&0.116{\scriptsize $\pm$0.0045}&0.222{\scriptsize $\pm$0.0033} \\

\midrule

\multicolumn{2}{c}{BE}& 0.364{\scriptsize $\pm$ 0.0073} & 0.227{\scriptsize $\pm$ 0.0020} &0.362{\scriptsize $\pm$ 0.0040}&0.241{\scriptsize $\pm$0.0031}&0.381{\scriptsize $\pm$0.0030}&0.266{\scriptsize $\pm$0.0022}&0.480{\scriptsize $\pm$0.0212}&0.316{\scriptsize $\pm$0.0100} \\

\midrule

\multicolumn{2}{c}{FR}& 0.354{\scriptsize $\pm$ 0.0047} &0.183{\scriptsize $\pm$ 0.0057} &0.357{\scriptsize $\pm$0.0026}&0.191{\scriptsize $\pm$0.0017}&0.390{\scriptsize $\pm$0.0074}&0.215{\scriptsize $\pm$0.0018}&0.469{\scriptsize $\pm$0.0197}&0.272{\scriptsize $\pm$0.0087} \\

\midrule

\multicolumn{2}{c}{DE}&0.277{\scriptsize $\pm$ 0.0097}& 0.321{\scriptsize $\pm$ 0.0030} &0.300{\scriptsize $\pm$0.0055}&0.338{\scriptsize $\pm$0.0016}&0.294{\scriptsize $\pm$0.0163}&0.340{\scriptsize $\pm$0.0094}&0.545{\scriptsize $\pm$0.0101}&0.481{\scriptsize $\pm$0.0056} \\

\midrule

\multicolumn{2}{c}{AVG}& 0.246{\scriptsize $\pm$ 0.0050}&0.219{\scriptsize $\pm$ 0.0028} &0.258{\scriptsize $\pm$ 0.0034}&0.236{\scriptsize $\pm$ 0.0020}&0.271{\scriptsize $\pm$ 0.00672}&0.252{\scriptsize $\pm$ 0.0035}&0.385{\scriptsize $\pm$ 0.0127}&0.323{\scriptsize $\pm$ 0.0067} \\
\bottomrule

\end{tabular}}
\end{table}

\section{User Study Details}
\label{user-study-details}
\begin{figure}[H]
    \centering
    \begin{adjustbox}{center,raise=0cm,margin*=0.1cm 0cm 0cm 0cm}
        \includegraphics[width=1\linewidth]{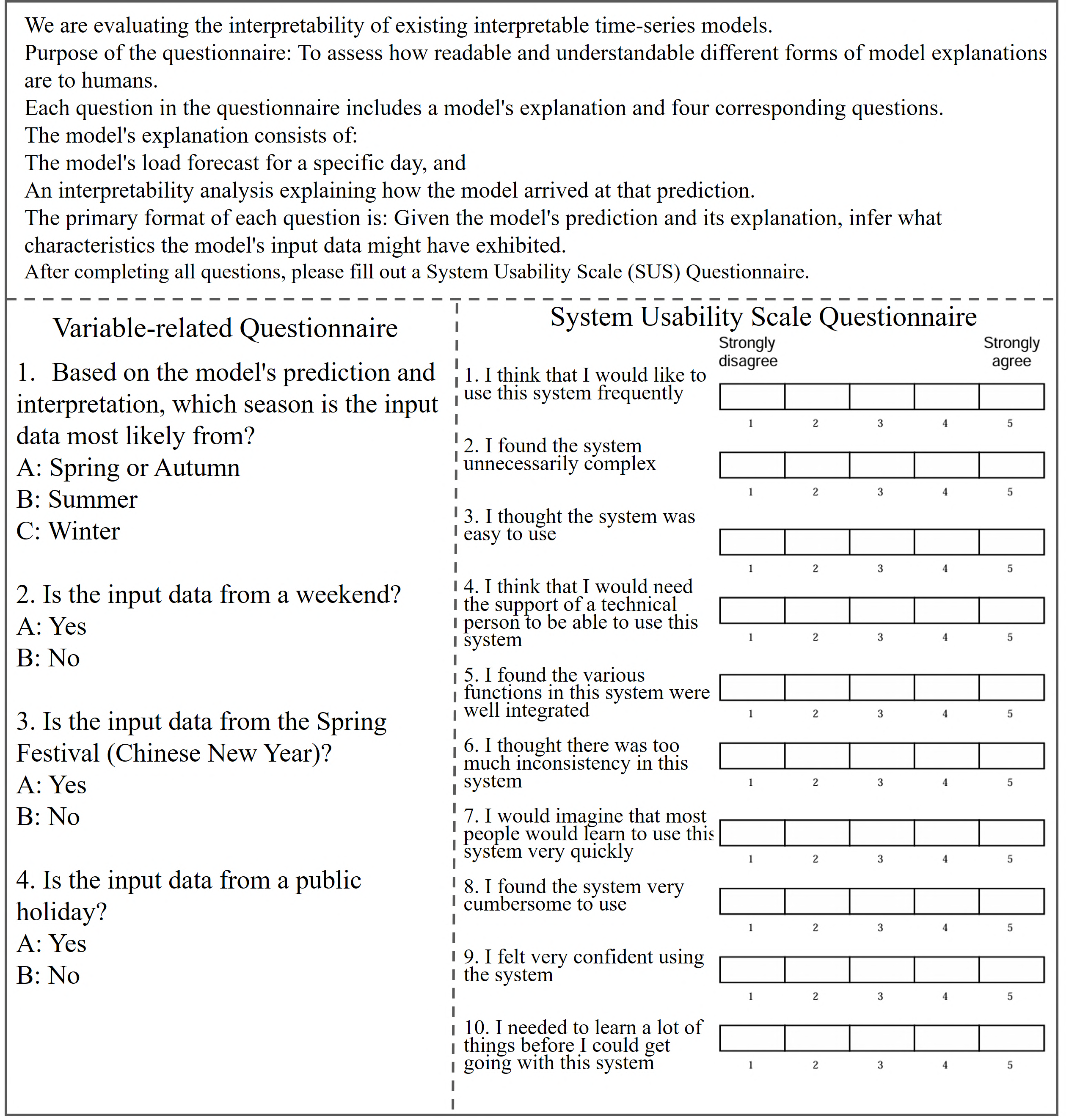}
    \end{adjustbox}
    \caption{User Study Questionnaire. \textbf{Bottom}: An overview of the questionnaire's purpose, question formats, and response instructions. \textbf{Left}: The Variable-related Questionnaire, where users are asked to infer input variables based on the provided explanations. Each question has a single correct answer, used to evaluate the accuracy of the system's explanations. \textbf{Right}: The System Usability Scale (SUS) Questionnaire, which captures users' subjective assessments to measure the usability of the current system.}
    \label{user-study-pic}
\end{figure}

We conducted the user study using a questionnaire format, the specific layout of which is shown in Figure~\ref{user-study-pic}. The questionnaire consists of two parts: Variable-related Questionnaire and System Usability Scale (SUS) Questionnaire. The first part is Variable-related Questionnaire, aims to evaluate the accuracy and comprehensibility of explanations provided by different explainable models. Participants are asked to perform causal inference on variables—such as holidays, seasons, Chinese New Year, and other related variables—based on the model’s predictions and its generated explanations. This section has unique, objective correct answers, and we measure the readability of the explanations by calculating the accuracy of participants’ responses. The second part is System Usability Scale (SUS) Questionnaire, designed to assess the usability of the explanation systems offering different levels of explainability. Participants are asked to subjectively rate the system based on their experience completing the first part~\citep{xu2024uncovering}. The SUS yields a single number representing a composite measure of the overall usability of the system under study. Note that scores for individual items are not meaningful on their own. To calculate the SUS score, first sum the score contributions from each item. Each item’s contribution ranges from 0 to 4. For items 1, 3, 5, 7, and 9, the contribution is the scale position minus 1. For items 2, 4, 6, 8, and 10, the contribution is 5 minus the scale position. Multiply the total sum of these contributions by 2.5 to obtain the overall SUS score, which ranges from 0 to 100. 

\section{Visual Comparison}
\label{Comparison}
\begin{figure}[H]
    \centering
    \begin{adjustbox}{center,raise=0cm,margin*=0cm 0cm 0cm 0cm}
        \includegraphics[width=1\linewidth]{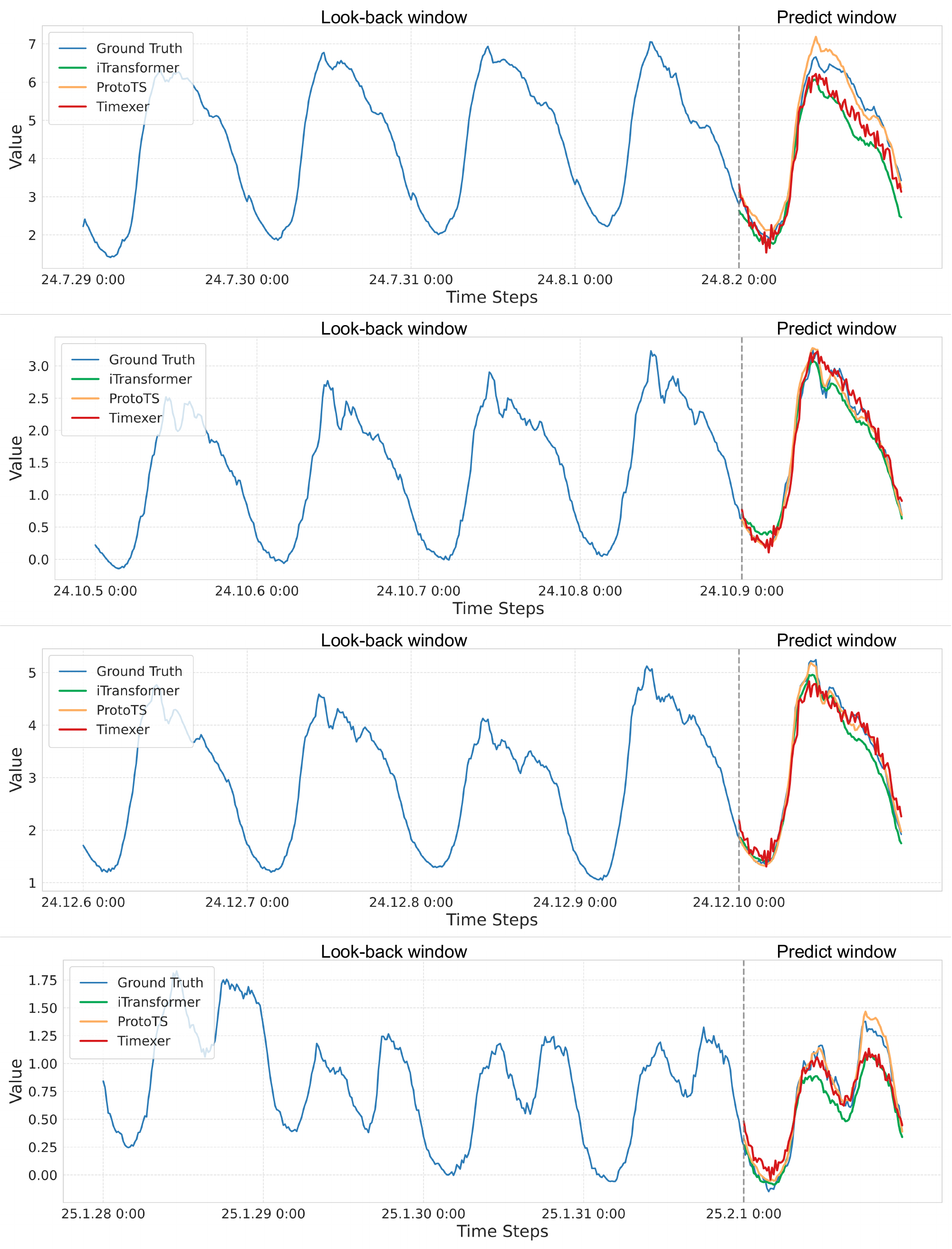}
    \end{adjustbox}
    \caption{Result Visualization: The prediction results for four sampled days, consistent with our experimental setting: Look-back window is 4 days and Predict window is 1 day, a day consists of 96 time steps. The horizontal axis represents time, and the vertical axis represents the current electricity load. The electricity load values have been normalized, so they are unitless and may include negative values.}
    \label{comparison}
\end{figure}

To intuitively illustrate the differences between ProtoTS and existing methods in predictions, we visualize the prediction results. The models chosen for comparison are iTransformer and Timexer, and the visualization is presented in Figure~\ref{comparison}. For the prediction on August $2^{nd}$, ProtoTS successfully captures the evening peak, whereas the other two models perform poorly. For the predictions on October $9^{th}$ and December $10^{th}$, although all three models capture the overall shape reasonably well, ProtoTS preserves significantly more fine-grained details. Regarding the prediction for February $1^{st}$, ProtoTS more accurately forecasts noon peak of the day.

\section{Discussion}
\label{discussion}
Although ProtoTS achieves high forecasting accuracy while providing multi-level, expert-steerable interpretability, it also has some potential limitations:
\begin{itemize}[nosep,leftmargin=1em,labelwidth=*,align=left]
    \item \textbf{Limited adaptability to non-periodic patterns}: ProtoTS relies on prototypical representations to capture typical temporal patterns. While this works well for seasonal or periodic data, its effectiveness diminishes when dealing with highly irregular, nonrepetitive time series. In such cases, more prototypes are needed to cover diverse patterns, increasing model complexity and reducing interpretability.
    \item \textbf{Dependence on domain expertise for refinement}: The interpretability of ProtoTS benefits from expert-guided prototype editing. However, this process requires sufficient domain knowledge and manual effort. In domains where expert resources are hard to access or knowledge is difficult to formalize, the usability and effectiveness of this feature may be limited~\citep{zhang2024distillation}.
    \item \textbf{Computational cost}: ProtoTS introduces additional computational overhead compared to transformer-based or MLP-based baselines (e.g., iTrans and TiDE). Under the same hardware and training configuration (optimizer, learning rate, sequence length, and batch size), ProtoTS shows higher per-iteration training time due to its multi-channel embedding design and hierarchical prototype modules, both of which require extra forward computations.
\end{itemize}

\begin{table}[t]
\centering
\begin{tabular}{lcccc}
\toprule
 & \textbf{ProtoTS} & \textbf{TimeXer} & \textbf{iTrans} & \textbf{TiDE} \\
\midrule
\textbf{ms/iter} & 40.8 & 19.5 & 12.5 & 11.1 \\
\bottomrule
\end{tabular}
\vspace{-0.5em}
\caption{Training time per iteration (ms).}
\label{tab:comp_cost}
\end{table}
\vspace{-1em}

\section{Multivariate Time Series Forecasting}

\label{multi}

In this section, we evaluate the performance of ProtoTS on standard multivariate time series forecasting tasks.
 
\textbf{Datasets}\quad We conduct comprehensive experiments on six benchmark datasets to evaluate the multivariate forecasting capability of ProtoTS. These datasets cover a variety of domains:
\begin{itemize}[nosep,leftmargin=1em,labelwidth=*,align=left]
    \item ETTh1 and ETTh2~\citep{zhou2021informer}: Hourly electricity load data from two different nodes in a power grid. They capture seasonal patterns and variable interactions across multiple features, serving as standard benchmarks for long-term forecasting.
    \item ETTm1 and ETTm2~\citep{zhou2021informer}: Minute-level electricity load datasets focusing on finer temporal granularity. They emphasize short-term dynamics and present challenges in modeling high-frequency fluctuations.
    \item Weather~\citep{zhou2021informer}: A multivariate weather dataset containing temperature, humidity, wind speed, and other meteorological variables collected from multiple stations.
    \item Exchange~\citep{lai2018modeling}: A financial time series dataset with daily exchange rates of eight foreign currencies. 
\end{itemize}

For all datasets, we follow standard experimental protocols, using a 96 look-back window and forecasting multiple future steps \{96, 192, 336, 720\} to evaluate both short-term accuracy and long-horizon stability.
 
\textbf{Baseline}\quad We compare ProtoTS against a broad set of baselines, including:
Transformer-based models (TimeXer~\citep{wang2024timexer}, iTransformer~\citep{liu2023itransformer}, PatchTST~\citep{nie2022time}, TimesNet~\citep{wu2022timesnet}, Crossformer~\citep{zhang2023crossformer}, Stationary~\citep{liu2022non}, Autoformer~\citep{wu2021autoformer}), MLP-based architectures (TiDE~\citep{das2023long}, DLinear~\citep{zeng2023transformers}, RLinear~\citep{li2023revisiting}, SCINet~\citep{liu2022scinet}).

\textbf{Results}\quad The results in Table~\ref{multi-results} indicate that ProtoTS achieves leading performance across most multivariate forecasting tasks. On structured datasets like ETTh1 and ETTh2, ProtoTS consistently outperforms all baselines at both short and long horizons, confirming its strength in modeling global temporal patterns. For high-frequency datasets such as ETTm1 and ETTm2, ProtoTS shows clear improvements in short-term forecasts, but its advantage narrows at longer horizons (e.g., 720 steps), where methods like PatchTST and TimesNet achieve comparable results. On the Weather dataset, ProtoTS maintains strong performance, though models like Crossformer remain competitive. For the challenging Exchange dataset, ProtoTS delivers robust forecasts but faces slight performance gaps against simpler models like DLinear at certain horizons. Overall, ProtoTS shows stable and strong performance across different tasks.

\clearpage
\begin{table}
  \caption{Descriptions of the exogenous variables used in the LOF dataset. The table includes each variable's data type, whether it is discrete or continuous, its physical unit (if applicable), and a brief description of its meaning and role in the dataset.}
  \label{LOF-covariates}
  \centering
  \scalebox{0.85}{
  \begin{tabular}{l c c c p{6.5cm}}
    \toprule
    \textbf{Name} & \textbf{Datatype} & \textbf{Type} & \textbf{Unit} & \textbf{Description} \\
    \midrule
    Solar year                       & int   & Discrete   & -         & gregorian calendar year                         \\
    \midrule
    Solar month                      & int   & Discrete   & -         & gregorian calendar month                        \\
    \midrule
    Solar day                        & int   & Discrete   & -         & gregorian calendar day                          \\
    \midrule
    Lunar year                       & int   & Discrete   & -         & chinese lunar calendar year                     \\
    \midrule
    Lunar month                      & int   & Discrete   & -         & chinese lunar calendar month                    \\
    \midrule
    Lunar day                        & int   & Discrete   & -         & chinese lunar calendar day                      \\
    \midrule
    Hour                             & int   & Discrete   & -         & hour of day                                     \\
    \midrule
    Minute                           & int   & Discrete   & -         & minute of hour                                  \\
    \midrule
    Skin temperature           & float & Continuous & °C        & the temperature of the surface of the Earth                   \\
    \midrule
    Surface pressure           & float & Continuous & hPa       & the temperature of sea water near the surface                 \\
    \midrule
    Surface sensible heat flux       & float & Continuous & Jm$^{-2}$  & the transfer of heat between the Earth's surface and the atmosphere through the effects of turbulent air motion                       \\
    \midrule
    Total cloud cover            & float & Continuous & \%        & the proportion of a grid box covered by cloud               \\
    \midrule
    Surface net solar radiation      & float & Continuous & Jm$^{-2}$  &  the amount of solar radiation that reaches a horizontal plane at the surface of the Earth minus the amount reflected by the Earth's surface             \\
    \midrule
    Total precipitation 4h lead    & float & Continuous & m         & the accumulated liquid and frozen water that falls to the Earth's surface    \\
    \midrule
    10 metre wind         & float & Continuous & ms$^{-1}$  & maximum 3 second wind at 10 metre height as defined                     \\
    \midrule
    Relative humidity            & float & Continuous & \%        & the water vapour pressure as a percentage of the value at which the air becomes saturated                  \\
    \midrule
    Is special workday               & bool  & Discrete   & -         & special workday added to make up for time lost during an adjusted holiday.          \\
    \midrule
    Is special holiday               & bool  & Discrete   & -         & public holiday mandated by law          \\
    \midrule
    Is holiday                       & bool  & Discrete   & -         & rest day, including weekends and public holidays             \\
    \midrule
    Is workday                       & bool  & Discrete   & -         & work day                           \\
    \midrule
    Day of week                      & int   & Discrete   & -         & monday=0, …, sunday=6                       \\
    \midrule
    Pricing                          & float & Continuous & -         & real‑time electricity price strategy            \\
    \bottomrule
  \end{tabular}}
\end{table}

\clearpage
\begin{table}\Large
\centering
\renewcommand{\arraystretch}{2}

\caption{Full results of multivariate time series forecasting task, where the input window is set to 96, and the forecast window is set to \{96, 192, 336, 720\}. Results are citepd from TimeXer~\citep{wang2024timexer}, otherwise reproduced. $1^{\mathrm{st}}$ denotes the first count.}
\label{multi-results}
\centering
\scalebox{0.34}{
\begin{tabular}{c c|c c|c c|c c|c c|c c|c c|c c|c c|c c|c c|c c|c c}
\toprule

\multicolumn{2}{c|}{Model} & \multicolumn{2}{c|}{ProtoTS} & \multicolumn{2}{c|}{TimeXer} & \multicolumn{2}{c|}{iTrans.} & \multicolumn{2}{c|}{RLinear} & \multicolumn{2}{c|}{PatchTST} & \multicolumn{2}{c|}{Cross.} & \multicolumn{2}{c|}{TiDE} & \multicolumn{2}{c|}{TimesNet}& \multicolumn{2}{c|}{DLinear} & \multicolumn{2}{c|}{SCINet} & \multicolumn{2}{c|}{Stationary}&\multicolumn{2}{c}{Auto.} \\

\midrule

\multicolumn{2}{c|}{Metric} & MSE & MAE & MSE & MAE & MSE & MAE & MSE & MAE & MSE & MAE & MSE & MAE & MSE & MAE & MSE & MAE & MSE & MAE & MSE & MAE& MSE & MAE & MSE & MAE \\
\midrule

\multirow{5}{*}{\rotatebox{90}{{\LARGE Etth1}}}&96 & \textbf{0.380} & \textbf{0.388} & 0.382 & 0.403 & 0.386 & 0.405 & 0.386 & 0.395 & 0.414 & 0.419 & 0.423 & 0.448 & 0.479 & 0.464 & 0.384 & 0.402 & 0.386 & 0.400 & 0.654 & 0.599 & 0.513 & 0.491 & 0.449 & 0.459 \\

& 192 &0.437&\textbf{0.423}& \textbf{0.429} & 0.435 & 0.441 & 0.436 & 0.437 & 0.424 & 0.460 & 0.445 & 0.471 & 0.474 & 0.525 & 0.492 &0.436 & 0.429 & 0.437 & 0.432 & 0.719 & 0.631 & 0.534 & 0.504 & 0.500 & 0.482 \\
& 336 &0.483&\textbf{0.440}&\textbf{0.468} & 0.448 & 0.487 & 0.458 & 0.479 & 0.446 & 0.501 & 0.466 & 0.570 & 0.546 & 0.565 & 0.515 & 0.491 & 0.469 & 0.481 & 0.459 &0.778 &0.659 & 0.588 & 0.535 & 0.521 & 0.496\\
& 720 &0.484&\textbf{0.459}&\textbf{0.469} & 0.461 & 0.503 & 0.491 & 0.481 & 0.470 & 0.500 & 0.488 & 0.653 & 0.621 & 0.594 & 0.558 &0.521 & 0.500 & 0.519 & 0.516 & 0.836 & 0.699 & 0.643 & 0.616 & 0.514 & 0.512\\

\cmidrule{3-26}

& AVG &0.445&\textbf{0.427}&\textbf{0.437}& 0.437 &0.454& 0.447 &0.446& 0.434& 0.469& 0.454& 0.529& 0.522& 0.541& 0.507& 0.458 &0.450 &0.456 &0.452 &0.747& 0.647 &0.570 &0.537& 0.496& 0.487\\

\midrule

\multirow{5}{*}{\rotatebox{90}{{\LARGE Etth2}}}& 96 & \textbf{0.284} &\textbf{0.337}& 0.286 & 0.338 & 0.297  &0.349 & 0.288 & 0.338 & 0.302  &0.348 & 0.745 & 0.584 & 0.400  &0.440 & 0.340 & 0.374 & 0.333  &0.387 & 0.707 & 0.621 & 0.476 & 0.458 & 0.346  &0.388 \\

&192 & 0.370 & \textbf{0.387}& \textbf{0.363} & 0.389 & 0.380 & 0.400 & 0.374 & 0.390 & 0.388 & 0.400 & 0.877 & 0.656 & 0.528 & 0.509 & 0.402 & 0.414 & 0.477 & 0.476 & 0.860 & 0.689 & 0.512 & 0.493 & 0.456 & 0.452 \\

&336 &\textbf{0.408}&\textbf{0.421}& 0.414 & 0.423 & 0.428 & 0.432 & 0.415 & 0.426 & 0.426 & 0.433 & 1.043 & 0.731 & 0.643 & 0.571 & 0.452 & 0.452 & 0.594 & 0.541 & 1.000 & 0.744 & 0.552 & 0.551 & 0.482 & 0.486 \\

&720 &\textbf{0.405}&\textbf{0.431}& 0.408 & 0.432 & 0.427 & 0.445 & 0.420 & 0.440 & 0.431 & 0.446 & 1.104 & 0.763 & 0.874 & 0.679 & 0.462 & 0.468 & 0.831 & 0.657 & 1.249 & 0.838 & 0.562 & 0.560 & 0.515 & 0.511 \\

\cmidrule{3-26}

&AVG &\textbf{0.366}&\textbf{0.394}&0.367&0.396& 0.383& 0.407& 0.374 &0.398 &0.387& 0.407& 0.942& 0.684& 0.611& 0.550& 0.414& 0.427& 0.559& 0.515& 0.954& 0.723& 0.526& 0.516 &0.450 &0.459\\

\midrule
\multirow{5}{*}{\rotatebox{90}{{\LARGE Ettm1}}} &96 &0.324&\textbf{0.347}& \textbf{0.318} & 0.356 & 0.334 & 0.368 & 0.355 & 0.376 & 0.329 & 0.367 & 0.404 & 0.426 & 0.364 & 0.387 & 0.338 & 0.375 & 0.345 & 0.372 & 0.418 & 0.438 & 0.386 & 0.398 & 0.505 & 0.475 \\

&192 &0.375&\textbf{0.369}&\textbf {0.362} & 0.383 & 0.387 & 0.391 & 0.391 & 0.392 & 0.367 & 0.385 & 0.450 & 0.451 & 0.398 & 0.404 & 0.374 & 0.387 & 0.380 & 0.389 & 0.426 & 0.441 & 0.459 & 0.444 & 0.553 & 0.496 \\

&336 &0.402&\textbf{0.389}& \textbf{0.395} & 0.407 & 0.426 & 0.420 & 0.424 & 0.415 & 0.399 & 0.410 & 0.532 & 0.515 & 0.428 & 0.425 & 0.410 & 0.411 & 0.413 & 0.413 & 0.445 & 0.459 & 0.495 & 0.464 & 0.621 & 0.537 \\

&720 &0.470&\textbf{0.426}&\textbf{0.452} & 0.441 & 0.491 & 0.459 & 0.487 & 0.450 & 0.454 & 0.439 & 0.666 & 0.589 & 0.487 & 0.461 & 0.478 & 0.450 & 0.474 & 0.453 & 0.595 & 0.550 & 0.585 & 0.516 & 0.671 & 0.561 \\

\cmidrule{3-26}

&AVG &0.392&\textbf{0.387}&0.382 &0.397& 0.407 &0.410 &0.414& 0.407 &0.387& 0.400& 0.513& 0.496& 0.419& 0.419& 0.400 &0.406 &0.403& 0.407& 0.485& 0.481& 0.481& 0.456& 0.588 &0.517 \\

\midrule

\multirow{5}{*}{\rotatebox{90}{{\LARGE Ettm2}}}&96&0.176&\textbf{0.254}&\textbf{0.171}& 0.256 & 0.180 & 0.264 & 0.182 & 0.265 & 0.175 & 0.259 & 0.287 & 0.366 & 0.207 & 0.305 & 0.187 & 0.267 & 0.193 & 0.292 & 0.286 & 0.377 & 0.192 & 0.274 & 0.255 & 0.339 \\

&192  &\textbf{0.237}&\textbf{0.295}&\textbf{0.237}& 0.299 & 0.250 & 0.309 & 0.246 & 0.304 & 0.241 & 0.302 & 0.414 & 0.492 &0.290&0.364&0.249&0.309&0.284&0.362&0.399&0.445&0.280&0.339&0.281&0.340\\

&336  &\textbf{0.295}&\textbf{0.332}& 0.296 & 0.338 & 0.311 & 0.348 & 0.307 & 0.342 & 0.305 & 0.343 & 0.597 & 0.542&0.377&0.422&0.321&0.351&0.369&0.427&0.637&0.591&0.334&0.361&0.339&0.372 \\

&720  &0.394&\textbf{0.390}&\textbf{0.392}& 0.394 & 0.412 & 0.407 &0.407& 0.398 & 0.402 & 0.400 & 1.730 & 1.042 & 0.558 & 0.524 & 0.408&0.403&0.554&0.522&0.960&0.735&0.417&0.413&0.433&0.432 \\

\cmidrule{3-26}
&AVG  &0.275&\textbf{0.312}&\textbf{0.274}&0.322 &0.288& 0.332 &0.286& 0.327& 0.281 &0.326& 0.757& 0.610& 0.358& 0.404& 0.291& 0.333& 0.350& 0.401& 0.571 &0.537 &0.306& 0.347& 0.327& 0.371 \\

\midrule

\multirow{5}{*}{\rotatebox{90}{{\LARGE weather}}} & 96&\textbf{0.155}&\textbf{0.197}& 0.157 & 0.205 & 0.174 & 0.214 & 0.192 & 0.232 & 0.177 & 0.218 & 0.158 & 0.230 &0.202&0.261&0.172&0.220&0.196&0.255&0.221&0.306&0.173&0.223&0.266&0.336\\

& 192 &\textbf{0.202}&\textbf{0.241}& 0.204 & 0.247 & 0.221 & 0.254 & 0.240 & 0.271 & 0.225 & 0.259 & 0.206 & 0.277&0.242&0.298&0.219&0.261&0.237&0.296&0.261&0.340&0.245&0.285&0.307&0.367 \\

& 336 &\textbf{0.259}&\textbf{0.285}& 0.261 & 0.290 & 0.278 & 0.296 & 0.292 & 0.307 & 0.278 & 0.297 & 0.272 & 0.335&0.287&0.335&0.280&0.306&0.283&0.335&0.309&0.378&0.321&0.338&0.359&0.395\\

& 720 &\textbf{0.337}&\textbf{0.333}& 0.340 & 0.341 & 0.358 & 0.349 & 0.364 & 0.353 & 0.354 & 0.348 & 0.398 & 0.418&0.351&0.386&0.365&0.359&0.345&0.381&0.377&0.427&0.414&0.410&0.419&0.428 \\

\cmidrule{3-26}
& AVG &\textbf{0.238}&\textbf{0.264}& 0.241&0.271&0.258&0.279&0.272& 0.291& 0.259 &0.281& 0.259& 0.315& 0.271& 0.320& 0.259& 0.287& 0.265& 0.317& 0.292& 0.363 &0.288 &0.314& 0.338& 0.382 \\

\midrule

\multirow{5}{*}{\rotatebox{90}{{\LARGE Exchange}}}&96&\textbf{0.085}&\textbf{0.205}&0.090&0.209&0.086&0.206&0.093&0.217&0.088&0.205&0.256&0.367&0.094&0.218&0.107&0.234&0.088&0.218&0.267&0.396&0.111&0.237&0.197&0.323\\

&192&\textbf{0.173}&\textbf{0.299}&0.190&0.309&0.177&\textbf{0.299}&0.184&0.307&0.176&0.299&0.470&0.509&0.184&0.307&0.226&0.344&0.176&0.315&0.351&0.459&0.219&0.335&0.300&0.369\\

&336&0.339&0.424&0.371&0.439&0.331&0.417&0.351&0.432&\textbf{0.301}&\textbf{0.397}&1.268&0.883&0.349&0.431&0.367&0.448&0.313&0.427&1.324&0.853&0.421&0.476&0.509&0.524 \\

&720&0.923&0.726&0.900&0.714&0.847&\textbf{0.691}&0.886&0.714&0.901&0.714&1.767&1.068&0.852&0.698&0.964&0.746&\textbf{0.839}&0.695&1.058&0.797&1.092&0.769&1.447&0.941
\\

\cmidrule{3-26}
&AVG&0.380&0.413&0.417&0.417&0.360 &\textbf{0.403} &0.378& 0.417& 0.367& 0.404& 0.940& 0.707& 0.370& 0.413 &0.416& 0.443& \textbf{0.354}& 0.414& 0.750& 0.626& 0.461& 0.454& 0.613& 0.539
\\

\midrule

& $1^{\mathrm{st}}$ Count& 14 & 27 & 13 & 0 &0 &3 &0 & 0& 1& 1& 0& 0 & 0 & 0 & 0 & 0 & 2& 0& 0& 0& 0& 0& 0& 0\\

\bottomrule

\end{tabular} }
\end{table}

\clearpage

\end{document}